\documentclass[10pt,journal,compsoc]{IEEEtran}

\usepackage[nocompress]{cite}
\usepackage[pdftex]{graphicx}
\usepackage{amsmath}
\usepackage{amssymb}
\usepackage{mathtools}
\usepackage{array}
\usepackage[caption=false,font=footnotesize,labelfont=sf,textfont=sf]{subfig}
\usepackage{dblfloatfix}
\usepackage{color}
\usepackage{algorithm}
\usepackage[noend]{algpseudocode}
\usepackage[switch]{lineno} 
\let\MYoriglatexcaption\caption
\renewcommand{\caption}[2][\relax]{\MYoriglatexcaption[#2]{#2}}
\usepackage{url}
\usepackage{csvsimple}
\usepackage{amsthm}
\usepackage{comment}

\usepackage{multirow}

\graphicspath{{./figure/segmentation/}{./figure/motion/}{./figure/denoise/}}
\DeclareGraphicsExtensions{.pdf,.jpeg,.png,.jpg}

\newcommand{\ud}{\,\mathrm{d}}
\newcommand{\dx}{\,\mathrm{d}x}

\newcommand{\dv}[1]{\mbox{div}\left( {#1} \right)}

\newcommand{\argmin}{\arg\!\min}
\newcommand{\argmax}{\arg\!\max}
\newcommand{\prox}{\textrm{prox}}

\def\f{{f}}
\def\u{{u}}
\def\v{{v}} 
\def\l{{\ell}}
\def\q{{q}}
\def\x{{x}}

\def\O{{\Omega}} 
\def\D{{\mathcal{D}}} 
\def\R{{\mathcal{R}}} 
\def\E{{\mathcal{E}}} 

\newtheorem{lemma}{Lemma}
\newtheorem{theorem}{Theorem}

\begin{document}

\title{Adaptive Regularization of Some Inverse Problems in Image Analysis}

\author{Byung-Woo~Hong,~\IEEEmembership{}
        Ja-Keoung~Koo,~\IEEEmembership{}
        Martin~Burger,~\IEEEmembership{}
        and~Stefano~Soatto~\IEEEmembership{}
\IEEEcompsocitemizethanks{
\IEEEcompsocthanksitem Byung-Woo Hong and Ja-Keoung Koo are with the Computer Science Department, Chung-Ang University, Korea. ({hong, jakeoung}@cau.ac.kr)
\IEEEcompsocthanksitem Martin Burger is with the Institute for Computational and Applied Mathematics, University of M\"unster,  Germany. (martin.burger@wwu.de)
\IEEEcompsocthanksitem Stefano Soatto is with the Computer Science Department, University of California Los Angeles, CA, USA (soatto@ucla.edu)}
}

\IEEEtitleabstractindextext{%
\begin{abstract}
We present an adaptive regularization scheme for optimizing composite energy functionals arising in image analysis problems. The scheme automatically trades off data fidelity and regularization depending on the current data fit during the iterative optimization, so that regularization is strongest initially, and wanes as data fidelity improves, with the weight of the regularizer being minimized at convergence.  We also introduce the use of a Huber loss function in both data fidelity and regularization terms, and present an efficient convex optimization algorithm based on the alternating direction method of multipliers (ADMM) using the equivalent relation between the Huber function and the proximal operator of the one-norm. We illustrate and validate our adaptive Huber-Huber model on synthetic and real images in segmentation, motion estimation, and denoising problems. 
\end{abstract}

}

\maketitle

\IEEEdisplaynontitleabstractindextext

%
\IEEEpeerreviewmaketitle


%
\IEEEraisesectionheading{\section{Introduction} \label{sec:introduction}}
\IEEEPARstart{I}{n} this paper we study problems of the composite form:
\begin{align}
\min_\u \D_\lambda(\u) + \R_\lambda(\u)
\label{eq:composite}
\end{align}
where $\u : \O \subset{\mathbb R}^2 \rightarrow \mathbb{R}^N; \x \mapsto \u(\x)$, and $\D$ is a data-dependent function and $\R$ is a regularization function:
\begin{align}
\D_\lambda(\u) &= \sum_{\x \in \O} \lambda(\u(x)) \, \rho(\u(x)), \label{eq:general:data}\\
\R_\lambda(\u) &= \sum_{\x \in \O} (1 - \lambda(\u(x))) \, \gamma(\u(x)), \label{eq:general:reg}
\end{align}
where $\rho(\u)$ and $\gamma(\u)$ are modulated by a function $\lambda(\u)$ that is allowed to vary in both space (the independent variable $\x$) and time (during the course of the optimization iteration).
%
%
Such an adaptive scheme generalizes both the classical Bayesian and Tikhonov regularization, with unique advantages that stem from the data-driven control of the amount of regularization.
%
%
%
Classically, one selects a model by picking a function(al) that measures {\em data fidelity}, which can be interpreted probabilistically as a log-likelihood, and one that measures {\em regularity}, which can be interpreted as a prior, with a parameter that trades off the two. 
%
%
%
%
\def\fh{77pt}
\begin{figure}
\centering
\includegraphics{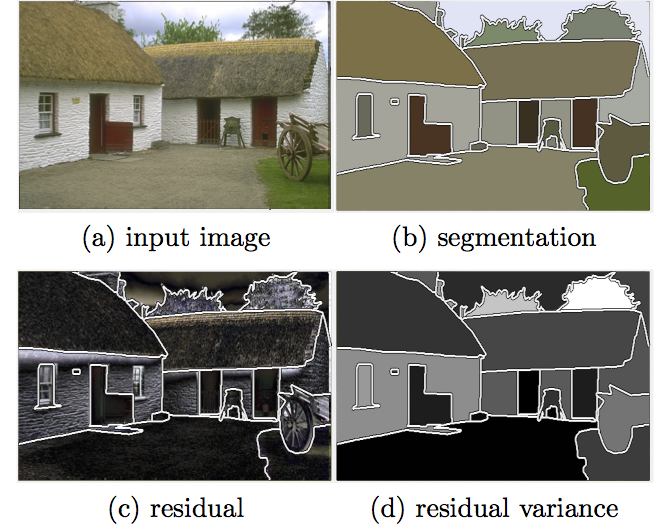}
\caption{Fixing the regularization parameter $\lambda$ {\em a-priori}, for instance in a segmentation task, results in excessive regularization in some regions, biasing the final solution (rounded corners, large $\lambda$), or otherwise is insufficient (jagged boundaries, small $\lambda$).
}
\label{fig:motivation}
\end{figure}
%
%
Typically, the trade-off between data fidelity and regularization assumed constant {\em both in space} (i.e., on the entire image domain $\Omega$)
{\em and in time}, i.e., during the entire course of the (typically iterative) optimization.
Neither is desirable. Consider for example a segmentation problem in Fig.~\ref{fig:motivation}: Panels (c) and (d) show the optimization residual and its variance, respectively, for each  region shown in (b), into which the image (a) is partitioned. Clearly, neither the residual, nor the variance (shown as a gray-level: bright is large, dark is small), are constant in space. 
This is also applicable to other imaging problems such as motion estimation (optical flow) and image restoration in which the local variance of the residual often varies in space and in the course of the optimization. 
Thus, {\em we need a spatially adapted regularization}, beyond static image features as studied in~\cite{lefkimmiatis2013convex,estellers2015adaptive}, or local intensity variations~\cite{dong2009multi,grasmair2009locally}. While regularization in these works is space-varying, the variation is tied to the image statistics, and therefore constant throughout the iteration. Instead, we propose a  {\em spatially-adaptive regularization scheme} that is a function of the residual, which changes during the iteration, yielding an automatically annealed schedule whereby the changes in the residual during the iterative optimization gradually guide the strength of the prior, adjusting it both in space, and in time/iteration.
In the modeling of conventional imaging problems, we present an efficient scheme that uses the Huber loss for both data fidelity and regularization, in a manner that includes standard models as a special case, within a convex optimization framework. While the Huber loss~\cite{huber1964robust} has been used before for regularization~\cite{chambolle2011first}, we use it both in the data and regularization terms.
Furthermore, to address the phenomenon of proliferation of multiple overlapping regions that plagues most multi-label segmentation schemes, we introduce a constraint that penalizes the common area of pairwise combinations of partitions. The classical constraints often used to this end are ineffective in a convex relaxation~\cite{chambolle2012convex}, which often leads to the need for user interaction~\cite{nieuwenhuis2014co,zemene:eccv:2016}.
We also present an annealing scheme between the forward and backward warpings in the computation of optical flow in order to better deal with large displacement.
Finally, we present an efficient convex optimization algorithm in the alternating direction method of multipliers (ADMM) framework~\cite{parikh2014proximal} with a variable splitting technique that enables us to effectively simplify this constraint~\cite{wang2008new}.
\subsection{Related work} \label{sec:relatedwork}
Regularization is commonly imposed to reduce the allowable space of solutions in several image analysis tasks which are formulated as ill-posed inverse problems. The associated parameters that modulate the strength of the regularizer are usually constant in both space and iteration and determined by grid search. In some cases, the parameters are tied to the data, for instance in image restoration where the noise variation has been used in~\cite{galatsanos1992methods} and the stability of the estimated parameter has been analyzed in~\cite{thompson1991study}. Framing the choice of parameter as model selection, cross-validation has also been used in~\cite{nguyen2001efficient}. 
Alternative approaches have been proposed based on the log-log plots of the norm of the residual and the regularization, called $L$-curve in~\cite{hansen1992analysis,mc2003direct}. However, the resulting methods are computationally expensive and often unstable when the variance of noise is small~\cite{vogel1996non}.
There have been other computationally expensive algorithms based on the truncated singular value decomposition~\cite{watzenig2004adaptive}, $U$-curve~\cite{krawczyk2007regularization}, and generalizations of the maximum likelihood estimate~\cite{wahba1985comparison} to determine global regularization. 
In the application of motion estimation, the regularization parameters have been inferred from the observed data in such a way that the joint probability of the gradient field and the velocity field is maximized~\cite{krajsek2006maximum}.
Other global approaches have used bilateral filtering~\cite{lee2010optical} and incorporated noise estimation~\cite{chantas2014variational} for the regularization of the estimated motion. 
There have been a number of spatially adaptive regularization schemes that incorporate the image gradient in the form of edge indicator function as a weighting factor for the regularization of optical flow~\cite{werlberger2009anisotropic,wedel2009structure} and image segmentation~\cite{grady2005multilabel,bresson2007fast}. 
A local variation of the image intensity within a fixed size window has been also used for modulating regularization~\cite{dong2009multi,grasmair2009locally}, and the regularization parameter has been chosen based on the variance~\cite{galatsanos1992methods}.
Non-local regularization has also been proposed for optical flow~\cite{krahenbuhl2012efficient,ranftl2014non}, image segmentation~\cite{bresson2008non,jung2011nonlocal}, and image restoration~\cite{perona1990scale} based on static image statistics. 
In contrast to static regularization, there have been dynamically adaptive methods that estimate the regularization parameter via a dynamic system in~\cite{kitchener2010adaptive} where the regularization is applied in a spatially global way. 
Other methods have been developed based on the Morozov's discrepancy principle~\cite{morozov1984regular} where the residual is bounded by the estimated noise in~\cite{aujol2006constrained,wen2012parameter}.
An anisotropic structure tensor has been used, based on total variation~\cite{lefkimmiatis2013convex,estellers2015adaptive} or generalized total variation~\cite{grasmair2010anisotropic,yan2015image}.
Most adaptive regularization algorithms have considered spatial statistics that are constant during the optimization iteration, irrespective of the residual. 

In conventional imaging tasks, the optimization has been widely performed based on discrete graph representations~\cite{grady2008reformulating,komodakis2011mrf} or continuous relaxation techniques~\cite{pock2010global,strekalovskiy2014convex} where Total Variation (TV) is used as a convex form of the regularization and its optimization is performed by a primal-dual algorithm. 
In minimizing TV, a functional lifting technique has been applied to the multi-label problem~\cite{pock2009algorithm,laude2016sublabel}.
Most convex relaxation approaches for multi-label problems have been based on TV regularization while different data fidelity terms have been used such as the  $L_1$~\cite{unger2012joint} or $L_2$ norms~\cite{brown2010convex}. Huber norms have been used for TV in order to avoid undesirable staircase effects~\cite{chambolle2011first}. Most multi-label models suffer from inaccurate or duplicate partitioned regions when using a large number of labels~\cite{chambolle2012convex}, which forces user interactions including bounding boxes~\cite{vicente2009joint}, contours~\cite{blake2004interactive}, scribbles~\cite{zemene:eccv:2016}, or points~\cite{bearman:eccv:2016}. 

%
\subsection{Summary of contributions} \label{sec:contribution}
Our primary contribution is to develop an adaptive regularization scheme in both space and time (optimization iteration) based on the local fit of observation to the model, which is measured by the data-driven statistics of the residual in the course of the optimization (Sect.~\ref{sect-motivation}).
We also introduce a composite energy functional that uses a robust Huber loss for both data fidelity and regularization, which are turned into the proximal operators of the $L_1$ norm via Moreau-Yosida regularization in a variety of imaging applications (Sect.~\ref{sec:application}).
In the image segmentation model, we propose a constraint on the mutual exclusivity of regions, which penalizes the common area of the pairwise combination of segmenting regions so that their assigned labels become more discriminative in particular with a number of region labels (Sect.~\ref{sec:segmentation:constrint}).
For the motion estimation, we introduce an annealing scheme that sequentially changes the degree of warping between forward and backward directions so that a large displacement can be effectively computed by considering both the forward and backward warpings (Sect.~\ref{sec:OpticalFlow:annealing}).
In order to demonstrate the robustness and effectiveness of our model, we perform quantitative and qualitative evaluation (Sect.~\ref{sec:experiment}).
\section{Adaptive Regularization} \label{sect-motivation}
In this section we motivate our approach to adaptive regularization by relating it formally to standard (Bayesian/Tikhonov) regularization. We indicate with $\f$ the data (for instance an image or video), $\u$ the object of interest (for instance the characteristic function of a partition of the image domain for segmentation, or the optical flow-field), and assume that we have a {\em model} in the form of a likelihood function $\l_\f(\u) \doteq p(\f|\u)$ and a prior $\q(\u) \doteq p(\u)$.
A Bayesian (maximum a-posteriori) criterion would then attempt to infer $\u$ by solving 
\[
\hat \u_{\rm \tiny bayes} = \arg\max_{\u} p(\u | \f) \propto \l(\u) \, \q(\u),
\]
where we have omitted the subscript $\f$ from $\l$. 
Often, these models are derived from an energy functional $\D(\u) = \sum_{\x \in \O} \rho(\u(\x))$, minimizing which is typically ill-posed, so Tikhonov regularization is imposed by selecting a functional $\R(\u) = \sum_{\x \in \O} \gamma(\u(\x))$, and minimizing $\D(\u) + \lambda \R(\u)$. Interpreting the data term $\D(\u)$ as a negative log-likelihood $\ell(\u(x)) \propto \exp(-\rho(\u(x)))$, and the regularizer $\R(\u)$ as a negative log-prior $\q(\u(\x)) \propto \exp(-\gamma(\u(\x)))$, we have 
\[
\hat \u_{\rm \tiny tikh} (\lambda) = \arg\max_{\u} \l(\u) \, \q^{(1 - \lambda)}(\u),
\]
where the multiplier $\lambda$ is a positive scalar parameter that controls the amount of regularization, and is fixed {\em a-priori}, and
\begin{align}
 \l(\u) &\propto \prod_{\x \in \O} \exp( - \rho(\u(\x))),\\
 \q(\u) &\propto \prod_{\x \in \O} \exp( - \gamma(\u(\x))).
\end{align}
The regularizer biases the final solution (which is a function of $\lambda$), establishing a trade-off between regularity (large $\lambda$) and fidelity (small $\lambda$). 
Instead of a fixed value, we can {\em change} $\lambda$ during the optimization procedure, so that the weight of the regularizer is maximal at first, and decreases subsequently, ideally to the point where it does not bias the final solution. For instance, we can choose 
\[
\hat \u_{\rm \tiny anneal} = \lim_{\lambda \rightarrow 1}\arg\max_\u \l(\u) \, \q^{(1 - \lambda)}(\u),
\]
where $\lambda \rightarrow 1$ according to some annealing schedule. Note that this does not have an interpretation in the Bayesian framework, and while it does not depend on the particular value of $\lambda$, it depends on the annealing schedule.
In our method, we instead choose a model of the form
\[
\hat \u_{\rm \tiny adapt}(\lambda) = \arg\max_u \l^{\lambda} \, \q^{(1 - \lambda)},
\]
where we omit the argument $u$ for ease of notation, and make $\lambda$ dependent on the solution $\u$ {\em pointwise}:
\[
\lambda(\x) \propto \l(\u(\x)).
\]
The rationale being that, when/where the solution is a poor fit of the data, the likelihood is small and therefore $\lambda$ is large and we impose heavy regularization, whereas when/where we have a perfect fit, the (normalized) likelihood approaches one and the effect of the regularizer is minimal. More importantly, $\lambda$ is different for each component $\x$ of $\u$, resulting in spatially-varying regularization, hence the name {\em adaptive regularity}. This model is adaptive in both space (component $\x$) and time (iteration). The resulting optimization is then
\begin{align}
\hat \u_{\rm \tiny adapt} = \arg\max_\u \sum_{\x \in \O} \l(\u) \log \l(\u) + (1 - \l(\u)) \log q(\u), \nonumber
\end{align}
where $x$ is omitted for $\u(\x)$ for ease of notation.
Often the above arises from the discretization of energy functionals under certain assumptions of conditional independence, as we describe next. 
\subsection{Assumptions} \label{sec:assumption}
In many cases of interest, the data $\f$ is distributed on a domain $\O$, and its values $\f(\x), \ \x \in \O$ can be modeled as samples from a stochastic process that has independent and identically distributed components {\em given} the value of $\u(\x)$:
\begin{align}
\l(\u) &= p(\f | \u) = \prod_{\x \in \O} p(\f(\x) | \u(\x)),\\
q(\u) &= p(\u) = \prod_{\x \in \O} p(\u(\x)).
\end{align}
%
%
Under these assumptions, the optimization above is equivalent to
\[
\hat \u_{\rm \tiny adapt} = \arg\max_\u \sum_{\x \in \O} \l(\u) \underbrace{\log \l(\u)}_{- \rho(u)}
+ (1 - \l(\u))\underbrace{\log \q(\u)}_{- \gamma(u)}.
\]
If we denote with $\rho(u(x)) = - \log(\l(\u(x)))$ the data-dependent energy, and $\gamma(u(x)) = - \log(\q(\u(\x)))$ the regularizing prior, we can also write the above as
\[
\hat \u_{\rm \tiny adapt} = \arg\min_\u \sum_{\x \in \O} e^{-\rho(\u)} \, \rho(\u) + (1 - e^{-\rho(\u)}) \, \gamma(\u).
\]
For mathematical convenience, we think of $\f$ as a continuous function $\f: {\mathbb R}^2 \rightarrow \mathbb{R}$ defined such that $\f(\x)$ coincides with the data on the lattice $\x \in \O \subset {\mathbb R}^2$, and consider functionals of the form
\[
\hat \u_{\rm \tiny adapt} = \arg\min_\u \int_\O e^{-\frac{\rho(\u)}{\beta}} \rho(\u) \dx + \int_\O \big(1 - e^{-\frac{\rho(\u)}{\beta}}\big) \gamma(\u) \dx,
\]
where $\beta$ is parameter corresponding to the variance of $\rho(\u)$.
%
%
More generally, we also allow for some amount of smoothing by a Gaussian kernel $G$, so we obtain models of the form:
\begin{align}
\E_{\lambda}(\u; \beta, G) &= \int_\O \lambda(\x) \, \rho(\u(\x)) + (1 - \lambda(\x)) \gamma(\u(\x)) \dx, \label{eq:general:energy}\\
\lambda(\x) &= \exp\left( -\frac{G * \rho(\u(\x))}{\beta}\right), \label{eq:general:lambda:kernel}
\end{align} 
with default choices $G = \delta$. The original cost functional \eqref{eq:composite} is then obtained by discretization.
%
%
%
%
\subsection{Analysis of Model} \label{sec:analysis}
A first general property that explains the behavior of the model in~\eqref{eq:general:energy} is the following:
\begin{lemma}
  Assume there exists $\u^*$ with $\rho(\u^*) \equiv 0$, then $\u^*$ is a fixed point of~\eqref{eq:general:energy}.
\end{lemma}
\begin{proof}
    Let $u^*$ satisfy $\rho(u^*)$, hence $\lambda \equiv 1$. Then 
    $$\E_{\lambda(\u^*)}(\u) = \int_\Omega \rho(\u(\x)) \dx, $$
    which is obviously minimized by $\u^*$. 
\end{proof}
The regularization is designed to manage non-convexity of the objective functional, but undesirable at convergence, where data fit is paramount.
We now provide a brief well-posedness analysis for the proposed model. 
For this sake, we consider the space to be minimized on $BV(\Omega)$ for $\Omega \subset \R^d$ a bounded domain. 
The mathematical definition of the regularization functional is then given by
$$ \mathcal{R}_\lambda(u) = \sup_{\varphi \in C_0^\infty(\Omega;\R^d), \Vert \varphi \Vert_\infty \leq 1} 
\int_\Omega u \nabla \cdot ( (1-\lambda) \varphi) \dx. $$
The problem we consider is then the minimization of
%
\begin{align} \label{eq:adaptiveentropy}
  \E_\lambda(u) &= \int_{\Omega} \lambda(\x)  \, \rho(\u(\x)) \dx + \mathcal{R}_{\lambda}(\u(\x))
\end{align}
with $\lambda(\x)$ as in \eqref{eq:general:lambda:kernel}.
%
%
For ease of mathematical presentation, we consider the minimization on the space of functions $\u$ of bounded variation with mean zero, which we  denote by $BV_0(\Omega)$.
In order to verify the existence of a solution for our model, it is natural to consider the fixed point map 
$ u \mapsto \lambda \mapsto u = \argmin \E_\lambda$, from which we derive the following result.
\begin{theorem} \label{thm:unique}
  Let $\beta > 0$ be sufficiently large. Let $G$ be bounded, integrable, and continuously differentiable with bounded and integrable gradient. Moreover, let $\rho$ be a continuous, nonnegative, convex functional, such that the minimizer of $\E_\lambda$ is unique for every $\lambda$.  Then there exists a fixed-point $\u \in BV_0(\O)$ for~\eqref{eq:adaptiveentropy}.
\end{theorem} 
The proof is provided in Appendix~\ref{sec:proof}. It is also noted that the uniqueness of $\E_\lambda$ in~\eqref{eq:adaptiveentropy} for fixed $\lambda$ is guaranteed if $\rho$ is strictly convex. 
%
%
\section{Application to Imaging Problems} \label{sec:application}
In this section, we present imaging models for segmentation, motion estimation and denoising problems based on the Huber-Huber model using our adaptive regularization scheme.
The problem of interest is cast as an energy minimization of the composite form \eqref{eq:composite}
%
%
where the relative weighting function $\lambda$ is defined by:
\begin{align} 
	\nu(x) &= \exp\left( -\frac{\rho(u(x)}{\beta} \right), \label{eq:composite:weight:nu}\\ 
	\lambda(x) &= \argmin_\lambda \frac{1}{2} \| \nu(x) - \lambda \|^2_2 + \alpha \| \lambda \|_1, \label{eq:composite:weight:lambda}
\end{align}
where $\beta > 0$ is a control parameter related to the variation of the residual $\rho(u)$, and $0 < \alpha < 1$ is a constant parameter to control the degree of sparsity in the weighting function $\lambda$ that is obtained by a solution of the Lasso problem~\cite{tibshirani1996regression}. 
The relative weight $\lambda$ between the data fidelity and the regularization is adaptively applied at each point $x$ depending on the residual $\rho(\u(x))$ determined by the local fit of data to the model.
The adaptive regularity scheme based on the weighting function $\lambda$ is designed so that regularization is stronger when the residual is large, equivalently $\nu$ is small, and weaker when the residual is small, equivalently $\nu$ is large, during the energy optimization process.
The range $0 < \nu \le 1$ with positive Lagrange multiplier $\alpha$ restricts the the weight $1 - \lambda$ to $[\alpha, 1)$ so that the regularization is imposed everywhere.     
In the definition of the data fidelity $\rho(\u)$ and the regularization $\gamma(\u)$, we use a robust Huber loss function $\phi_{\mu}$ with a threshold parameter $\mu > 0$~\cite{huber1964robust}: 
\begin{align} \label{eq:HuberFunction}
\phi_{\mu}(x) =
   \begin{cases}
   \frac{1}{2 \mu} x^2 & : |x| \le \mu,\\
   |x| - \frac{\mu}{2} & : |x| > \mu.
   \end{cases}
\end{align}
The advantage of using the Huber loss in comparison to the $L_2$ norm is that geometric features such as edges are better preserved while it has continuous derivatives in contrast to the $L_1$ norm that is not differentiable leading to staircase artifacts.
In addition, the Huber loss enables efficient convex optimization due to its equivalence to the proximal operator of $L_1$ norm, which will be discussed in Sect.~\ref{sec:optimization}.
%
%
\subsection{Image Segmentation via Adaptive Regularization} \label{sec:segmentation}
%
\subsubsection{Segmentation based on Huber-Huber model} \label{sec:segmentation:general}
Let $f : \Omega \rightarrow \mathbb{R}$ be a real valued\footnote{Vector-valued images can also be handled, but we consider scalar for ease of exposition.} image with domain $\Omega \subset {\mathbb R}^2$.
Segmentation aims to divide the domain $\Omega$ into a set of $n$ pairwise disjoint regions $\Omega_i$ where $\Omega = \cup_{i=1}^n \Omega_i$ and $\Omega_i \cap \Omega_j = \varnothing$ if $i \neq j$.  
The partitioning is represented by a labeling function $l : \Omega \rightarrow \Lambda$ where $\Lambda$ denotes a set of labels with $| \Lambda | = n$. The labeling function $l(x)$ assigns a label to each point $x \in \Omega$ such that $\Omega_i = \{ x \, | \, l(x) = i \}$.  
Each region $\Omega_i$ is indicated by the characteristic function $\chi_i : \Omega \rightarrow \{ 0, 1 \}$:
\begin{align} \label{eq:CharacteristicFunction}
\chi_i(x) = 
   \begin{cases}
   1 & : l(x) = i,\\
   0 & : l(x) \neq i.
   \end{cases}
\end{align}
Segmentation of an image $f(x)$ is obtained by seeking for regions $\{ \Omega_i \}$ that minimize an energy functional with respect to a set of characteristic functions $\{ \chi_i \}$:
\begin{align}
	\sum_{i \in \Lambda} \left\{ \D_{\lambda_i}(\chi_i) + \R_{\lambda_i}(\chi_i) \right\}, 
	\quad \sum_{i \in \Lambda} \chi_i(x) = 1. \label{eq:NonConvexEnergy}
\end{align}
%
For the data fidelity, we use a simple piecewise constant image model with an additional noise process: $f(x \, | \, \Omega_i) = c_i + \xi_i(x)$ with $c_i \in \mathbb{R}$ where $\xi_i$ is assumed to follow a bimodal distribution where its center follows a Gaussian distribution and its tails follow a Laplace distribution leading to the Huber loss function $\phi_\mu$: 
\begin{align}
    \D_{\lambda_i}(\chi_i, c_i) &= \int_\Omega \lambda_i(x) \, \rho(\chi_i(x), c_i) \dx, \label{eq:nonconvex:data}\\
    \rho(\chi_i(x), c_i) &= \phi_\mu(f(x) - c_i) \, \chi_i(x), \label{eq:nonconvex:rho}
\end{align}
where the weighting function $\lambda_i$ for label $i$ is determined based on the residual $\rho(\chi_i, c_i)$ as defined in~\eqref{eq:composite:weight:nu} and~\eqref{eq:composite:weight:lambda}.
For the regularization, we use a standard length penalty for each region $\Omega_i$:
\begin{align}
    \R_{\lambda_i}(\chi_i) &= \int_\Omega (1 - \lambda_i(x)) \, \gamma(\chi_i(x)), \label{eq:nonconvex:regular}\\
    \gamma(\chi_i(x)) &= \phi_\eta( \nabla \chi_i(x) ), \label{eq:nonconvex:gamma}
\end{align}
where $\eta > 0$ is a threshold for the Huber function $\phi_\eta$.
The energy formulation in~\eqref{eq:NonConvexEnergy} in terms of the characteristic function $\chi_i$ is non-convex due to its integer constraint $\chi_i \in \{0, 1\}$.  
We derive the convex form of the energy functional using classical convex relaxation methods~\cite{chambolle2011first,chambolle2012convex} where $\chi_i$ is replaced by a continuous function $\u_i \in BV(\Omega)$ of bounded variation and its integer constraint $\chi_i \in \{0, 1\}$ is relaxed into the convex set $\u_i \in [0, 1]$:
%
%
\begin{align}
	\sum_{i \in \Lambda} \left\{ \D_{\lambda_i}(\u_i, c_i) + \R_{\lambda_i}(\u_i) \right\},
	\quad \sum_{i \in \Lambda} \u_i(x) = 1, \label{eq:convex:energy}
\end{align}
where $\u_i : \Omega \rightarrow [0, 1]$ is a smooth function, and the data fidelity $\rho(u_i, c_i)$ and the regularization $\gamma(u_i)$ are defined by:
%
%
%
\begin{align} 
  \rho(\u_i(x), c_i) &= \phi_\mu( f(x) - c_i ) \, \u_i(x), \label{eq:label:huber:rho}\\
  \gamma(\u_i(x)) &= \phi_\eta( \nabla \u_i(x) ). \label{eq:label:huber:gamma}
\end{align}
%
The weighting function $\lambda_i$ is determined based on the residual $\rho(u_i, c_i)$ as defined in~\eqref{eq:composite:weight:nu} and~\eqref{eq:composite:weight:lambda} imposing a higher regularization to the partitioning function in which mismatch between the model and the observation occurs. In contrast, a lower regularization is imposed for the regions where the local observation fits the model.  
%
\subsubsection{Mutually Exclusive Region Constraint} \label{sec:segmentation:constrint}
The partitioning regions are constrained to be disjoint, however the condition $\sum_{i \in \Lambda} u_i(x) = 1$ in~\eqref{eq:convex:energy} along is ineffective in enforcing this constraint, in particular with a large number of labels~\cite{chambolle2012convex}.
Thus, we introduce a novel constraint to penalize the common area of each pair of combinations in regions $\Omega_i$ in such a way that $\sum_{i \neq j} u_i u_j$ is minimized for all $i, j \in \Lambda$. Then, we add it to the energy in~\eqref{eq:convex:energy} and arrive at the following:
\begin{align} \label{eq:ConvexEnergyAreaConstraint}
	\sum_{i \in \Lambda} & \bigg\{ \D_{\lambda_i}(u_i, c_i) + \R_{\lambda_i}(u_i) + \int_\Omega \tau \Big( \sum_{i \neq j} u_j(x) \Big) u_i(x) \dx \bigg\} \nonumber\\
&\textrm{subject to } u_i(x) \in [0, 1], \quad \sum_{i \in \Lambda} u_i(x) = 1,
\end{align}
where $\tau > 0$ is a weighting parameter for the mutual exclusivity constraint. 
The desired segmentation results are obtained by the optimal set of partitioning functions $u_i$:
\begin{align} \label{eq:ComputeLabel}
l(x) = \argmax_{i \in \Lambda} u_i(x).
\end{align}
%
%
%
\subsection{Optical Flow via Adaptive Regularization} \label{sec:OpticalFlow}
%
\subsubsection{Optical Flow based on Huber-Huber model} \label{sec:OpticalFlow:general}
Let $I : \Omega \times \mathbb{R} \rightarrow \mathbb{R}$ be a sequence of images $f(x; t)$ taken at space $x \in \Omega$ and time $t \in \mathbb{R}$. The optical flow problem aims to compute the velocity field $\u : \Omega \rightarrow \mathbb{R}^2$ that accounts for the motion between a pair of images $f_1(x) \coloneqq f(x; t)$ and $f_2(x) \coloneqq f(x; t+\Delta t)$.
The desired velocity field $\u$ is obtained by minimizing an energy functional that consists of the data fidelity and the regularization.
%
%
For the data fidelity, we consider an optical flow model based on the brightness consistency assumption~\cite{horn1981determining} with an additional noise process $\xi$ as follows:
\begin{align} \label{eq:consistency}
	f_2(x) = f_1(x + \u(x)) + \xi(x),
\end{align}
where $\u$ is an infinitesimal deformation of the image domain.
For ease of computation, we can apply a first-order Taylor series expansion with respect to a prior velocity field solution $u_0 : \Omega \rightarrow \mathbb{R}^2$ to linearize the first term:
\begin{align} \label{eq:taylor}
	&f_1(\x + \u(\x)) = f_1(\x + \u_0(\x) + \u(\x) - \u_0(\x)) \nonumber\\
	& = f_1(\x + \u_0(\x)) + \nabla f_1(\x + \u_0(\x)) \cdot (\u(\x) - \u_0(\x)),
\end{align}
where $\nabla f_1$ denotes the spatial gradient of image $f_1$, and the superscript notation for the transpose of the gradient is omitted from $\nabla f_1$ for simplicity. Then, the linearization of the brightness consistency condition in~\eqref{eq:consistency} and~\eqref{eq:taylor} leads to the following optical flow equation:
\begin{align} \label{eq:opticalflow}
	f_t(x) - \nabla f_1(x + u_0(x)) \cdot (u(x) - u_0(x)) = \xi(x),
\end{align}
where $f_t = f_2 - f_1$ denotes the temporal derivative of $f$.
We assume that the noise process $\xi$ follows a bimodal distribution leading to the Huber loss function $\phi_\mu$ with a threshold $\mu > 0$:
\begin{align}
	\D_\lambda(u) &= \int_\Omega \lambda(x) \, \rho(u(x)) \dx, \label{eq:motion:data}\\
	\rho(u) = \phi_\mu(f_t(x) &- \nabla f_1(x + u_0(x)) \cdot (u(x) - u_0(x))), \label{eq:motion:rho}
\end{align}
%
where the weighting function $\lambda$ is determined based on the residual $\rho(u)$ as defined in~\eqref{eq:composite:weight:nu} and~\eqref{eq:composite:weight:lambda}.
For the regularization, we use a standard smoothness term using the Huber loss function $\phi_\eta$ with a threshold $\eta > 0$:
%
\begin{align}
	\R_\lambda(u) &= \int_\Omega (1 - \lambda(x)) \, \gamma(u(x)) \dx, \label{eq:motion:reg}\\
	\gamma(u(x)) &= \phi_\eta(\nabla u_1(x)) + \phi_\eta(\nabla u_2(x)), \label{eq:motion:gamma}
\end{align}
%
where $\u(x) = (\u_1(x), \u_2(x))$ are the components of the velocity field. 
Note that the regularizer is necessary in regions where either the aperture problem is manifest ({\em e.g.,} in homogeneous regions, so $u$ is not unique) or at occlusion regions, where $u$ is not defined. However, regularization should not affect the solution where $u$ is well defined. 
%
\subsubsection{Annealing in Warping} \label{sec:OpticalFlow:annealing}
In computing $u$, we consider both forward and backward deformations of the domain with a control parameter $\tau \in \mathbb{R}$:
\begin{align}
	f_2(x - (1 - \tau) u(x)) = f_1(x + \tau u(x)) + \xi(x),	
\end{align}
where $\tau$ is to consider the degree of warping between the forward and the backward directions.
We introduce a simple annealing process for the control parameter $\tau$ the value of which gradually changes from $0.5$ to $1$ in the optimization procedure.
In considering the annealing of the warping direction, the data fidelity is modified as follows:
\begin{align}
	\D_{\lambda, \tau} (u) &= \int_\Omega \lambda(x) \rho_\tau(u(x)) \dx, \label{eq:motion:data:warp}\\
	\rho_\tau(u(x)) &= \phi_\mu( f_t(x) - ((1 - \tau) \nabla f_2 + \tau \nabla f_1) \, u(x) ), \label{eq:motion:data:warp:rho}
\end{align}
where $\lambda$ is determined based on the residual $\rho_\tau(\u(x))$, and the initial velocity field $\u_0$ is omitted for ease of presentation.
Then, the energy functional reads:
\begin{align} \label{eq:energy:motion:warp}
	\lim_{\tau \rightarrow 1} \argmin_{\u} \D_{\lambda, \tau} (\u) + \R_\lambda(\u),
\end{align}
where the initial value is $\tau = 0.5$ that gives the symmetric form of the energy, and $\tau \rightarrow 1$ increases subsequently according to an annealing process.
One simple example of the annealing process is based on the optimization iteration. 
%
%
\subsection{Denoising via Adaptive Regularization} \label{sec:denoising}
%
\subsubsection{Denoising based on Huber-Huber model} \label{sec:denoising:problem}
Let $f : \Omega \rightarrow \mathbb{R}$ be an observation and $u : \Omega \rightarrow \mathbb{R}$ be the reconstruction based on the additive noise assumption $f = u + \xi$ where $\xi$ denotes the noise process that is assumed to follow a bimodal distribution.
The desired reconstruction is obtained again by minimizing the energy functional 
%
%
where the data fidelity is defined by the Huber function $\phi_\mu$ with a threshold $\mu > 0$:
\begin{align}
	\D_\lambda(u) & = \int_\Omega \lambda(x) \, \rho(u(x)) \dx, \label{eq:energy:denoise:data}\\
	\rho(u(x)) & = \phi_\mu( f(x) - u(x) ), \label{eq:energy:denoise:rho} 
\end{align}
and the regularization is defined by the Huber function $\phi_\eta$ with a threshold $\eta > 0$:
\begin{align}
	\R_\lambda(u) & = \int_\Omega (1 - \lambda(x)) \, \gamma(u(x)) \dx, \label{eq:energy:denoise:reg}\\
	\gamma(u(x)) & = \phi_\eta( \nabla u(x) ). \label{eq:energy:denoise:gamma}
\end{align}
The weighting function $\lambda$ is determined based on the residual $\rho(u)$ as defined by~\eqref{eq:composite:weight:nu} and~\eqref{eq:composite:weight:lambda}.
%
\section{Energy Optimization} \label{sec:optimization}
In this section, we present optimization algorithms for the considered imaging problems in the framework of alternating direction method of multipliers (ADMM)~\cite{boyd2011distributed,parikh2014proximal} where the objective functional is of the following essential form:
\begin{align}
\min_\u \int_\O \lambda(\x) \, \rho(\u(\x)) \dx + \int_\O (1 - \lambda(\x)) \, \gamma(\u(\x)) \dx,
\end{align}
where $\lambda$ is determined by~\eqref{eq:composite:weight:nu} and~\eqref{eq:composite:weight:lambda}.
We initially modify the energy functional by the variable splitting that introduces a new variable $\v$ such that $\u = \v$:
%
\begin{align}
\int_\O & \lambda(\x) \, \rho(\u(\x)) \dx + \int_\O (1 - \lambda(\x)) \, \gamma(\v(\x)) \dx, \textrm{ with } u = v,\nonumber 
\end{align}
which leads to the following unconstrained augmented Lagrangian:
\begin{align}
\int_\O & \lambda \, \rho(\u) \dx + \int_\O (1 - \lambda) \, \gamma(\v) \dx + \frac{\theta}{2} \| u - v + w \|_2^2, \label{eq:general:lagrangian:unconstrained}
\end{align}
where $\theta > 0$ is a scalar augmentation parameter, and $w$ is a dual variable for the equality constraint $u = v$. 
In our imaging problems, the data fidelity $\rho(\u)$ and the regularization $\gamma(\v)$ are defined by the Huber function $\phi_\mu$, which can be efficiently optimized by Moreau-Yosida regularization of a non-smooth function $| \cdot |$ as given by~\cite{moreau1965proximite,opac-b1133911}:
\begin{align} \label{eq:MoreauYosida}
\phi_{\mu}(x) = \inf_r \left\{ | r | + \frac{1}{2 \mu} (x - r)^2 \right\} = \prox_{\mu g}(x),
\end{align}
where $r$ is an auxiliary variable to be minimized, and the proximal operator $\prox_{\mu g}(x)$ is associated with a convex function $g(x) = \| x \|_1$.
The solution of the proximal operator of the $L_1$ norm $\prox_{\mu g}(x)$ can be obtained by the soft shrinkage operator $\mathcal{T}(x | \mu)$ defined by~\cite{boyd2004convex}:
\begin{align}
\mathcal{T} (x \, | \, \mu) &= 
\begin{cases}
x - \mu & : x > \mu\\
0 & : \| x \|_1 \le \mu\\
x + \mu & : x < - \mu\\
\end{cases}
\label{eq:shrink}
\end{align}
The data fidelity $\rho(\u)$ and the regularization $\gamma(\v)$ in~\eqref{eq:general:lagrangian:unconstrained} can be replaced with $\rho(\u, r)$ and $\gamma(\v, z)$, respectively, by Moreau-Yosida regularization where $r$ and $z$ are the auxiliary variables to be minimized.
Then, we have the following general form of the energy functional $\mathcal{L}$:
\begin{align}
\mathcal{L}&(\u, \v, w, r, z) = \int_\O \lambda(\x) \, \rho(\u(\x), r) \dx \nonumber\\
&+ \int_\O (1 - \lambda(\x)) \, \gamma(\v(\x), z) \dx + \frac{\theta}{2} \| u - v + w \|_2^2, \label{eq:general:lagrangian:huber}
\end{align}
where $\lambda$ is determined by $\rho(\u, r)$.
The general optimization algorithm proceeds to minimize the augmented Lagrangian $\mathcal{L}$ in~\eqref{eq:general:lagrangian:huber} by applying a gradient descent scheme with respect to the variables $\u, \v, r, z$ and a gradient ascent scheme the dual variable $w$ followed by the update of the weighting function $\lambda$.
%
%
\begin{algorithm}[htb]
\caption{The ADMM updates for minimizing~\eqref{eq:general:lagrangian:huber}}
\label{alg:admm:general}
\begin{algorithmic}
\State 
\begin{flalign}
	r^{k+1} & \coloneqq \argmin_{r} \rho( u^k, r ) & \label{general:step:r}\\
	z^{k+1} & \coloneqq \argmin_z \gamma(v^k, z) & \label{general:step:z}\\
	u^{k+1} & \coloneqq \argmin_{u} \int_\Omega \lambda^{k+1} \, \rho(u, r^{k+1}) \dx & \nonumber\\
		& \quad + \frac{\theta}{2} \| u - v^{k+1} + w^k \|_2^2  & \label{general:step:u}\\
	v^{k+1} & \coloneqq \argmin_{v} \int_\Omega (1 - \lambda^{k+1}) \, \gamma(v, z^{k+1}) \dx & \nonumber\\
		& \quad + \frac{\theta}{2} \| u^k - v + w^k \|_2^2  & \label{general:step:v}\\
	w^{k+1} & \coloneqq w^{k} + u^{k+1} - v^{k+1} & \label{general:step:w}\\
	\nu^{k+1} & \coloneqq \exp\left( - \frac{\rho(u^{k+1}, r^{k+1})}{\beta} \right) & \label{general:step:nu}\\
	\lambda^{k+1} & \coloneqq \argmin_{\lambda} \frac{1}{2} \| \nu^{k+1} - \lambda \|_2^2 + \alpha \| \lambda \|_1 & \label{general:step:l}
\end{flalign}
\end{algorithmic}
\end{algorithm}
%
%
The alternating optimization steps for minimizing $\mathcal{L}$ in~\eqref{eq:general:lagrangian:huber} are presented in Algorithm~\ref{alg:admm:general}, where $k$ is the iteration counter. 
The more detailed optimization steps for each imaging problem will be presented in the following sections. The technical details regarding the optimality conditions and the optimal solutions are provided in Appendix~\ref{sec:optimality}.
%
%
\subsection{Optimization for Image Segmentation} \label{sec:optimization:segmentation}
The energy functional for the segmentation problem in~\eqref{eq:ConvexEnergyAreaConstraint} is minimized with respect to a set of partitioning functions $\{ u_i \}$ and intensity estimates $\{ c_i \}$ in an expectation-maximization (EM) framework. 
We apply the variable splitting $u_i = v_i$ to the energy functional in~\eqref{eq:ConvexEnergyAreaConstraint} as presented in~\eqref{eq:general:lagrangian:unconstrained}, and simplify the constraints as follows:
\begin{align} \label{eq:ConvexEnergySplit}
&\sum_{i \in \Lambda} \bigg\{ \int_\Omega \lambda_i \, \rho(u_i, c_i) + \tau \Big( \sum_{i \neq j} u_j \Big) u_i \ud x \nonumber\\
&+ \int_\Omega (1 - \lambda_i) \, \gamma( v_i ) \ud x + \frac{\theta}{2} \| u_i - v_i + w_i \|_2^2 \bigg\}, \nonumber\\
&\textrm{subject to } u_i(x) \ge 0, \, \sum_{i \in \Lambda} v_i(x) = 1, \, \forall x \in \Omega,
\end{align}
where $w_i$ is a dual variable for the equality constraint $u_i = v_i$ that allows to decompose the original constraints $u_i \in [0, 1]$ and $\sum_i u_i = 1$ into the simpler constraints $u_i \ge 0$ and $\sum_i v_i = 1$.
The data fidelity $\rho(u_i, c_i)$ in~\eqref{eq:label:huber:rho} and the regularization $\gamma(v_i)$ in~\eqref{eq:label:huber:gamma} can be replaced with the regularized forms $\rho(u_i, c_i, r_i)$ and $\gamma(v_i, z_i)$, respectively:
\begin{align} 
	\rho(u_i, c_i, r_i) &= \inf_{r_i} \left\{ \left( | r_i | + \frac{1}{2 \mu} (f - c_i - r_i)^2 \right) u_i \right\}, \label{eq:DataFidelityMoreauYosida}\\
	\gamma(v_i, z_i) &= \inf_{z_i} \left\{ \| z_i \|_1 + \frac{1}{2 \eta} \| \nabla v_i - z_i \|^2_2 \right\}, \label{eq:RegularizationMoreauYosida}
\end{align}
where $r_i$ and $z_i$ are the auxiliary variables to be minimized.
The constraints on $u_i$ and $v_i$ in~\eqref{eq:ConvexEnergySplit} can be represented by the indicator function $\delta_A(x)$ of a set $A$ defined by:
\begin{align} \label{eq:DeltaFunction}
\delta_A(x) &= 
   \begin{cases}
   0 & : x \in A,\\
   \infty & : x \notin A.\\
   \end{cases}
\end{align}
The constraint $u_i \ge 0$ is given by $\delta_A(u_i)$ where $A = \{ x | x \ge 0 \}$, and the constraint $\sum_i v_i = 1$ is given by $\delta_B( \{ v_i \} )$ where $B = \left\{ \{x_i\} | \sum_i x_i = 1 \right\}$.
The regularized forms of the data fidelity and the regularization, and the indicator functions for the constraints lead to the following unconstrained augmented Lagrangian $\mathcal{L}_i$ for label $i$:
\begin{align} \label{eq:EnergyUnconstrainedLabel}
&\mathcal{L}_i = \int_\Omega \lambda_i \, \rho(u_i, c_i, r_i) + \tau \Big( \sum_{i \neq j} u_j \Big) u_i \ud x + \delta_A(u_i) \nonumber\\
&+ \int_\Omega (1 - \lambda_i) \, \gamma( \nabla v_i, z_i ) \ud x + \frac{\theta}{2} \| u_i - v_i + w_i \|_2^2,
\end{align}
and the final energy functional $\mathcal{L}$ reads:
\begin{align} \label{eq:EnergyUnconstrained}
\mathcal{L}(\{u_i, v_i, w_i, c_i, r_i, z_i\}) = \sum_{i \in \Lambda} \mathcal{L}_i + \delta_B(\{v_i\}).
\end{align}
The optimal set of partitioning functions $\{ u_i \}$ is obtained by minimizing the energy functional $\mathcal{L}$.
The optimization proceeds to minimize the augmented Lagrangian for each label $\mathcal{L}_i$ in~\eqref{eq:EnergyUnconstrainedLabel} by following Algorithm~\ref{alg:admm:general}.
The obtained intermediate solutions $u_i$ and $v_i$ are projected onto the sets $A$ and $B$, respectively.
The algorithm is repeated until convergence from a given initialization for labeling function $l(x)$.
The technical details regarding the optimality conditions and the optimal solutions are provided in Appendix~\ref{sec:optimality:segmentation}.
%

%
%
%
\subsection{Optimization for Optical Flow} \label{sec:optimization:motion}
The energy functional for the optical flow in~\eqref{eq:energy:motion:warp} is minimized with respect to the velocity field $u$.
The intermediate solution of $u$ is iteratively used as the initial prior solution $u_0$, and the image warping is applied accordingly.
We apply the variable splitting $u = v$ introducing a new variable $v = (v_1, v_2)$ to the energy functional in~\eqref{eq:energy:motion:warp} as follows:
\begin{align} 
	\int_\Omega & \lambda \, \rho_\tau(u) \dx + \int_\Omega (1 - \lambda) \, \gamma(v) \dx + \frac{\theta}{2} \sum_{i = 1}^2 \| u_i - v_i + w_i \|_2^2, \label{eq:energy:motion:split}
\end{align}
where $w = (w_1, w_2)$ is a dual variable for the equality constraint $u = v$.
The data fidelity $\rho_\tau(u)$ in~\eqref{eq:motion:data:warp:rho} and the regularization $\gamma(v)$ in~\eqref{eq:motion:gamma} can be replaced with the regularized forms $\rho_\tau(u, r)$ and $\gamma(v, z)$:
\begin{align}
	\rho_\tau(u, r) &= \inf_{r} \left\{ | r | + \frac{1}{2 \mu} ( f_t - (\nabla f_1 + \tau \nabla f_2) \, u - r )^2 \right\}, \label{eq:motion:data:warp:rho:MoreauYosida}\\
	\gamma(v, z) &= \inf_{z} \left\{ \sum_{i = 1}^2 \left( \| z_i \|_1 + \frac{1}{2 \eta} \| \nabla v_i - z_i \|_2^2 \right) \right\}, \label{eq:motion:gamma:MoreauYosida}
\end{align}
where $r$ and $z = (z_1, z_2)$ are the auxiliary variables to be minimized.
Then, the augmented Lagrangian $\mathcal{L}$ reads:
\begin{align}
	& \mathcal{L} (u, v, w, r, z) = \int_\Omega \lambda(x) \, \rho_\tau(u, r) \dx \nonumber\\
  & + \int_\Omega (1 - \lambda(x)) \, \gamma(v, z) \dx + \frac{\theta}{2} \sum_{i = 1}^2 \| u_i - v_i + w_i \|_2^2. \label{eq:energy:motion:lagrangian} 
\end{align}
The desired velocity field $u = (u_1, u_2)$ is obtained by minimizing $\mathcal{L}$ in~\eqref{eq:energy:motion:lagrangian} and we follow the optimization procedure in Algorithm~\ref{alg:admm:general}.
For the control parameter $\tau$ for the warping annealing, we use a simple scheme that increases from $0.5$ to $1$ by a given step size $\Delta \tau \in \mathbb{R}$ at each iteration.
The technical details regarding the optimality conditions and the optimal solutions are provided in Appendix~\ref{sec:optimality:motion}.
%
%
%
%
\subsection{Optimization for Denoising} \label{sec:optimization:denoise}
The objective functional to optimize for the denoising problem reads:
\begin{align} 
  \int_\Omega \lambda \, \rho(u, r) \dx + \int_\Omega (1 - \lambda) \, \gamma(v, z) \dx + \frac{\theta}{2} \| u - v + w \|_2^2, \label{eq:energy:denoise:split}
\end{align}
where the data fidelity $\rho(u, r)$ and the regularization $\gamma(v, z)$ are defined by:
\begin{align}
	\rho(u, r) & = \inf_r \left\{ |r| + \frac{1}{2 \mu} (f - u - r)^2 \right\}, \label{eq:denoise:data:MoreauYosida}\\
	\gamma(v, z) & = \inf_z \left\{ \|z\| + \frac{1}{2 \eta} \|\nabla v - z\|_2^2 \right\}, \label{eq:denoise:reg:MoreauYosida}
\end{align}
where $r$ and $z$ are the auxiliary variables.
%
%
Then, the augmented Lagrangian $\mathcal{L}$ reads: 
\begin{align} 
	\mathcal{L} & (u, v, w, r, z) = \int_\Omega \lambda(x) \, \rho(u, r) \dx \nonumber\\
  & + \int_\Omega (1 - \lambda(x)) \, \gamma(v, z) \dx + \frac{\theta}{2} \| u - v + w \|_2^2. \label{eq:energy:denoise:lagrangian}
\end{align}
%
%
We follow the optimization steps in Algorithm~\ref{alg:admm:general} until convergence from the initial condition $u = f$.
The technical details regarding the optimality conditions and the optimal solutions are provided in Appendix~\ref{sec:optimality:denoise}.
%
%
%
%
%
\section{Experimental Results} \label{sec:experiment}
In this section, we demonstrate the robustness and effectiveness of our proposed adaptive regularization scheme in the application of segmentation, motion estimation and denoising.
The numerical experiments aim to present the relative advantage of using our proposed adaptive regularization scheme over the conventional static one.
We employ a classical imaging model and compare the performance of the given model with the modified algorithm that replaces the original regularization with our proposed one. 
We also demonstrate the advantage in using our Huber-Huber model.
Note that the adaptive regularization can be integrated into more sophisticated models by merely replacing their regularization parameter with our adaptive weighting function based on the residual of the model under consideration.
%
%
\subsection{Multi-Label Segmentation} \label{sec:experiment:segmentation}
In the experiments, we use the images in the Berkeley segmentation dataset~\cite{arbelaez2011contour} and simple yet illustrative synthetic ones.
Note that we use a random initialization for the initial labeling function for all the algorithms throughout the experiments.
\subsubsection{Robustness of the Huber-Huber ($\mathrm H^2$) model} \label{sec:experiment:segmentation:HuberHuber}
%
%
\def\fh{87pt}
\begin{figure}[htb]
\centering
\begin{tabular}{c@{}c}
\includegraphics[totalheight=\fh]{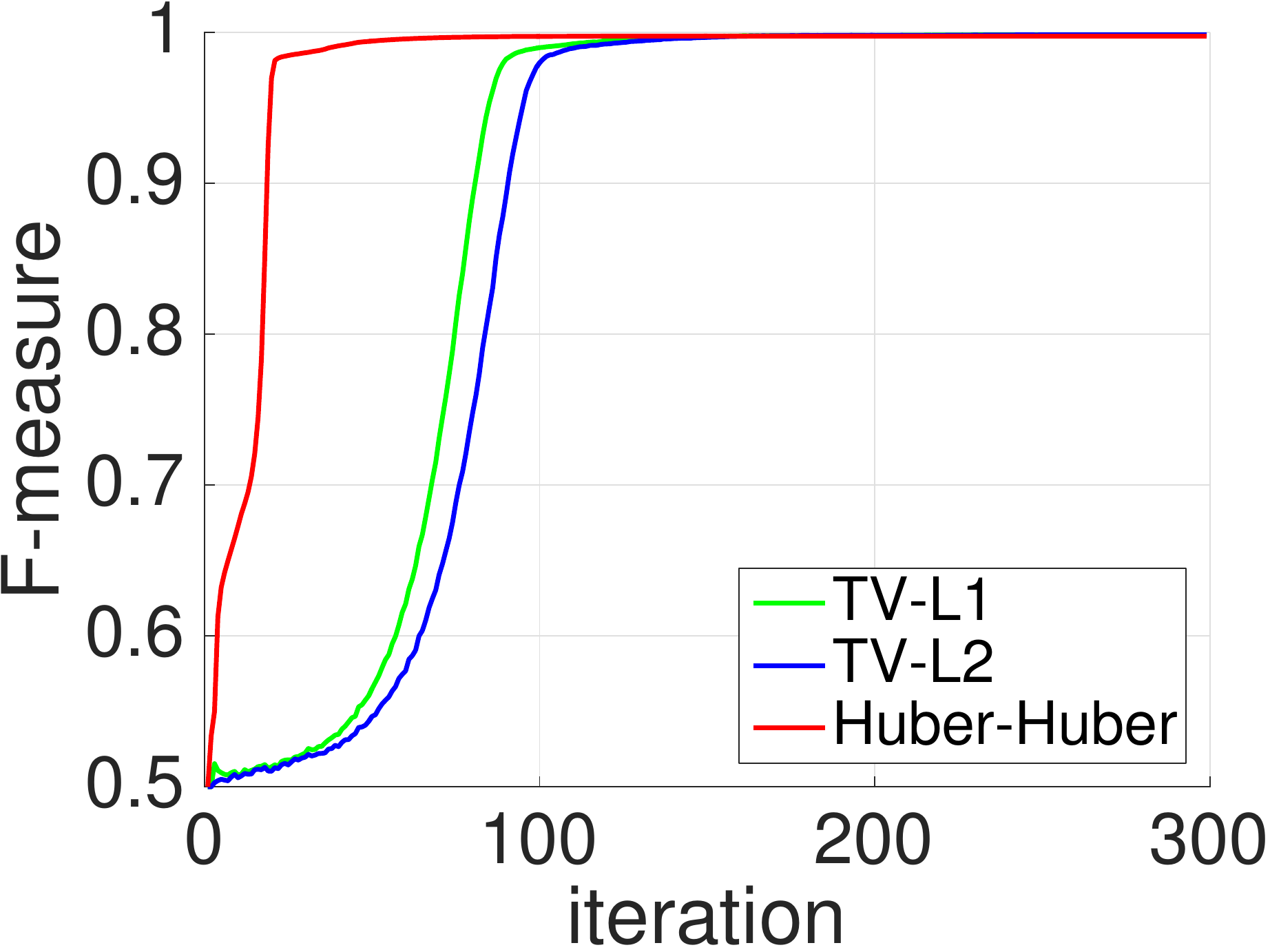} &
\includegraphics[totalheight=\fh]{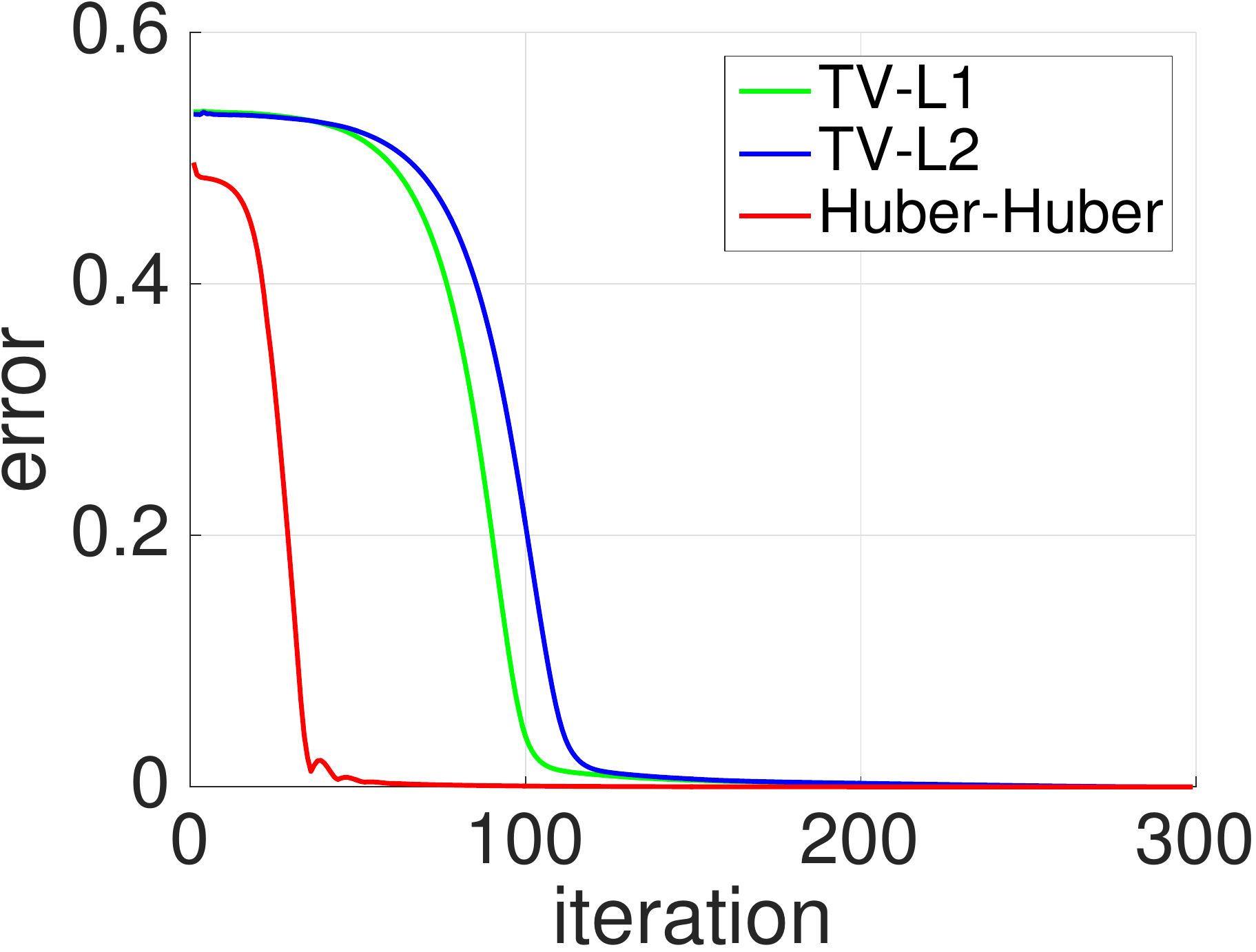}\\
(a) Accuracy & (b) Error
\end{tabular}
\caption{Quantitative Comparison of the different energy models using the F-measure (left) and residual (right) as a function of the number of iterations. We compare the popular TV-$L_1$ and TV-$L_2$ approaches to our ${\rm H}^2$ model, which is more accurate (left) and also converges faster (right) for a bi-partitioning problem on the images suited for the bi-partitioning image model.}
\label{fig:compare:model:accuracy}
\end{figure}
%
%
%
We consider a bi-partitioning image model in order to effectively demonstrate the robustness of our Huber-Huber model in comparison to TV-$L_1$ and TV-$L_2$ ignoring the effect of the constraint on the common area of the pairwise region combinations. 
For the optimization, we apply the primal-dual algorithm for TV-$L_1$ and TV-$L_2$ models.
It is shown that our Huber-Huber model yields better results faster as shown in Fig.~\ref{fig:compare:model:accuracy} where (a) F-measure and (b) error are presented for each iteration. 
The parameters for each algorithm are chosen fairly in such a way that the accuracy and convergence rate are optimized. 
%
\subsubsection{Effectiveness of Mutually Exclusive Constraint} \label{sec:experiment:segmentation:constraint}
%
%
\setlength{\fboxsep}{0pt}
\setlength{\fboxrule}{0pt}
\def\fh{52pt}
\def\fw{70pt}
\def\sp{7pt}
\begin{figure*}[hbt]
\centering
\begin{tabular}{c@{}c@{}c@{}c@{}c@{}c@{}c@{}c}
\fbox{\parbox[b][\fh][c]{\fw}{\# of regions (5)\\\# of labels (4)}} &
\includegraphics[trim={20pt 20pt 20pt 20pt},clip,totalheight=\fh]{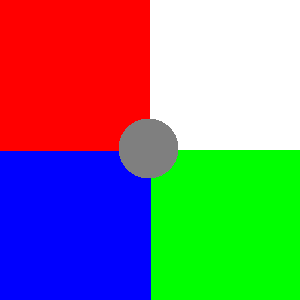} \hspace{\sp} & 
\includegraphics[trim={20pt 20pt 20pt 20pt},clip,totalheight=\fh]{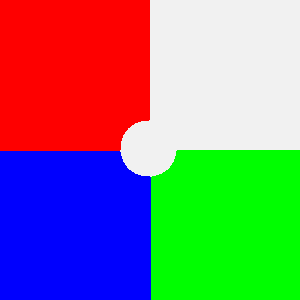} \hspace{\sp} &  
\includegraphics[trim={20pt 20pt 20pt 20pt},clip,totalheight=\fh]{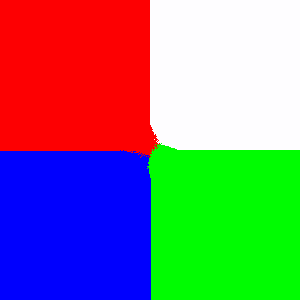} \hspace{\sp} &  
\includegraphics[trim={20pt 20pt 20pt 20pt},clip,totalheight=\fh]{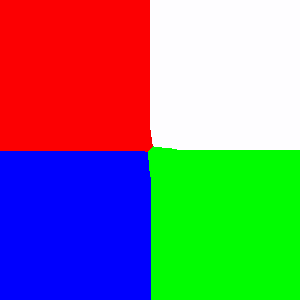} \hspace{\sp} &  
\includegraphics[trim={20pt 20pt 20pt 20pt},clip,totalheight=\fh]{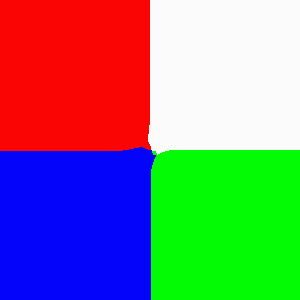} \hspace{\sp} &  
\includegraphics[trim={20pt 20pt 20pt 20pt},clip,totalheight=\fh]{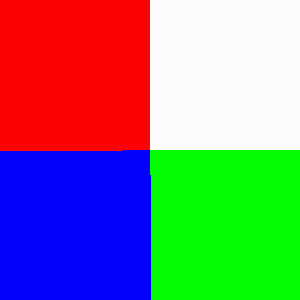} \hspace{\sp} & 
\includegraphics[trim={20pt 20pt 20pt 20pt},clip,totalheight=\fh]{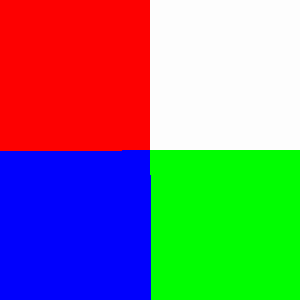} \\
\fbox{\parbox[b][\fh][c]{\fw}{\# of regions (7)\\\# of labels (6)}} &
\includegraphics[trim={20pt 20pt 20pt 20pt},clip,totalheight=\fh]{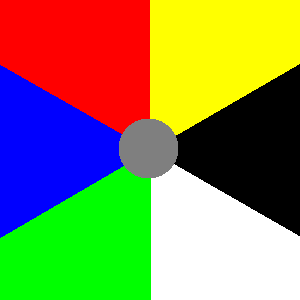} \hspace{\sp} &
\includegraphics[trim={20pt 20pt 20pt 20pt},clip,totalheight=\fh]{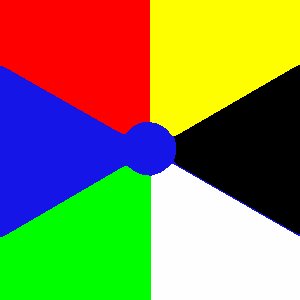} \hspace{\sp} &
\includegraphics[trim={20pt 20pt 20pt 20pt},clip,totalheight=\fh]{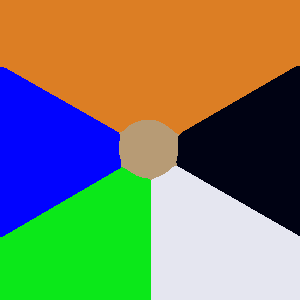} \hspace{\sp} &
\includegraphics[trim={20pt 20pt 20pt 20pt},clip,totalheight=\fh]{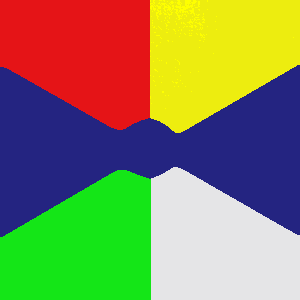} \hspace{\sp} &
\includegraphics[trim={20pt 20pt 20pt 20pt},clip,totalheight=\fh]{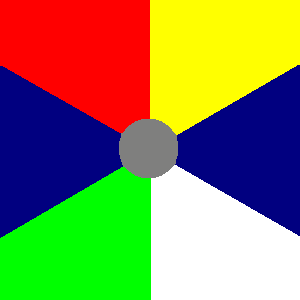} \hspace{\sp} &
\includegraphics[trim={20pt 20pt 20pt 20pt},clip,totalheight=\fh]{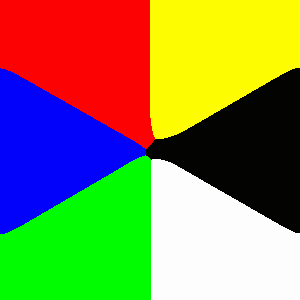} \hspace{\sp} &
\includegraphics[trim={20pt 20pt 20pt 20pt},clip,totalheight=\fh]{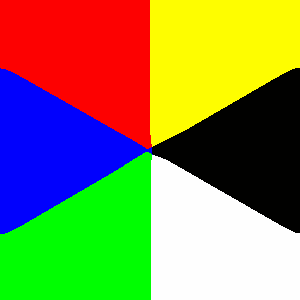} \\
\fbox{\parbox[b][\fh][c]{\fw}{\# of regions (9)\\\# of labels (8)}} &
\includegraphics[trim={20pt 20pt 20pt 20pt},clip,totalheight=\fh]{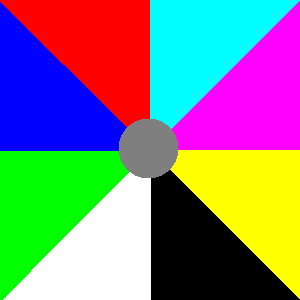} \hspace{\sp} &
\includegraphics[trim={20pt 20pt 20pt 20pt},clip,totalheight=\fh]{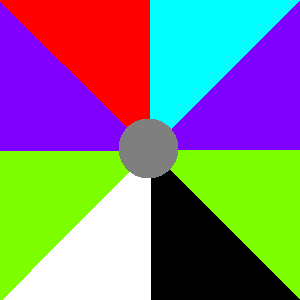} \hspace{\sp} &
\includegraphics[trim={20pt 20pt 20pt 20pt},clip,totalheight=\fh]{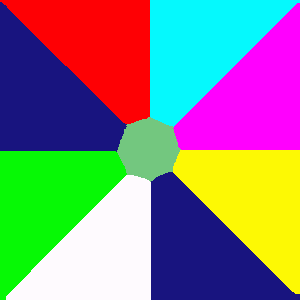} \hspace{\sp} &
\includegraphics[trim={20pt 20pt 20pt 20pt},clip,totalheight=\fh]{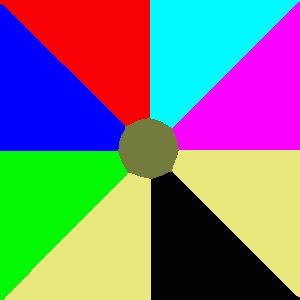} \hspace{\sp} &
\includegraphics[trim={20pt 20pt 20pt 20pt},clip,totalheight=\fh]{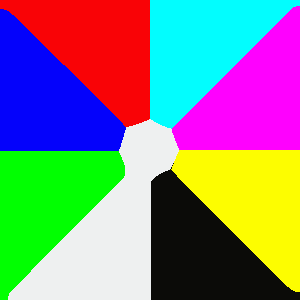} \hspace{\sp} &
\includegraphics[trim={20pt 20pt 20pt 20pt},clip,totalheight=\fh]{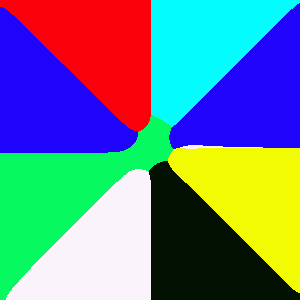} \hspace{\sp} &
\includegraphics[trim={20pt 20pt 20pt 20pt},clip,totalheight=\fh]{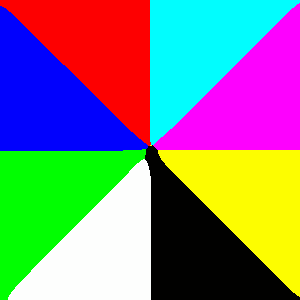} \\
& (a) Input \hspace{\sp} & (b) FL~\cite{sundaramoorthi2014fast} \hspace{\sp} & (c) TV~\cite{zach2008fast} \hspace{\sp} & (d) VTV~\cite{lellmann:continuous:siam:2011} \hspace{\sp} & (e) PC~\cite{chambolle2012convex} \hspace{\sp} & (f) our $\rm{H}^2$ \hspace{\sp} & (g) our full $\rm{H}^2$
\end{tabular}
\caption{Qualitative comparison for the {\em junction test} with increasing number of regions (5, 7, 9 from top to bottom). The number of labels is fixed at one-minus the true one (4, 6, 8 respectively), forcing the algorithm to fill in one of the regions. This test is reflective of the ability of the prior to capture the structure of the image in the presence of missing data. We compare multiple models, as indicated in the legend, all of which degrade with the number of regions; ours shows consistently better performance, as indicated by more regular in-painting (g).}
\label{fig:compare:model:constraint}
\end{figure*}
%
%
%
We qualitatively compare the segmentation results with different number of labels on the classical {\em junction test} (cf. Fig.~12 of~\cite{chambolle2011first} or Fig.~5-8 of~\cite{chambolle2012convex}), whereby the number of labels is fixed to one-less than the number of regions in the input image. The algorithm is then forced to {\em ``inpaint''} the central disc with labels of surrounding regions. 
The segmentation results on the junction prototype images with different number of regions are shown in Fig.~\ref{fig:compare:model:constraint} where the input junction images have 5 (top), 7 (middle), 9 (bottom) regions as shown in (a). 
We compare (f) our $\rm{H}^2$ model without the mutual exclusivity constraint and (g) our full $\rm{H}^2$ model with the constraint to the algorithms including: (b) fast-label (FL)~\cite{sundaramoorthi2014fast}, (c) convex relaxation based on Total Variation using the primal-dual (TV)~\cite{zach2008fast}, (d) vectorial Total Variation using the Dogulas-Rachford (VTV)~\cite{lellmann:continuous:siam:2011}, (e) paired calibration (PC)~\cite{chambolle2012convex}.
This experiment is particularly designed to demonstrate a need for the constraint of the mutual exclusivity, thus the input images are made to be suited for a precise piecewise constant model so that the underlying image model of the algorithm under comparison is relevant. 
The illustrative results shown in Fig.~\ref{fig:compare:model:constraint} indicate that the most algorithms degrades as the number of regions increases (top-to-bottom), while our algorithm yields consistently better results.

\subsubsection{Effectiveness of Adaptive Regularity} \label{sec:experiment:segmentation:adaptive}
%
%
%
%
\def\fh{75pt}
\def\sp{10pt}
\begin{figure}[!h]
\centering
\footnotesize
\begin{tabular}{c@{}c@{}c}
\includegraphics[totalheight=\fh]{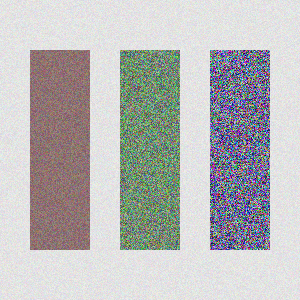} & 
\includegraphics[totalheight=\fh]{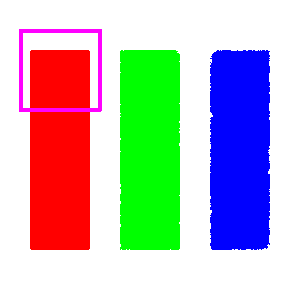} &
\includegraphics[totalheight=\fh]{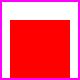} \\
(a) Input & (b) Ours (adaptive) & (c) Zoom in of (b)\\
\includegraphics[totalheight=\fh]{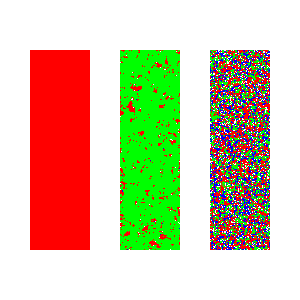} &  
\includegraphics[totalheight=\fh]{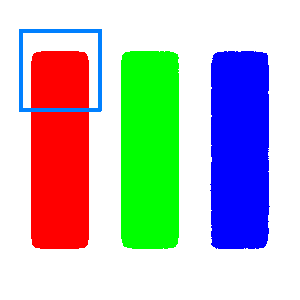} &
\includegraphics[totalheight=\fh]{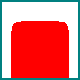} \\
(d) Small (global) & (e) Large (global) & (f) Zoom in of (e)
\end{tabular}
\caption{Qualitative comparison illustrating the pitfalls of a constant, non-adaptive, regularizer and its bias on the final solution. The image (a) has three regions with sharp boundaries/corners, and varying amount of spatial variability. Using a small amount of regularization (d) (large $\lambda$) yields sharp boundaries, but irregular partition into three regions (red, green and blue), with different labels dispersed throughout the center and right rectangles. Using a large regularizer weight (e) (small $\lambda$) yields homogeneous regions, but the boundaries are blurred out and the corners rounded, even in the left region (red). Our method (b) is driven by the data where possible (left region, sharp boundaries) and let the regularizer weigh-in where the data is more uncertain (right region, rounded boundaries).}
\label{fig:compare:regularity}
\end{figure}
%
%
We empirically demonstrate the effectiveness of our proposed adaptive regularization using an illustrative synthetic image with four regions, each exhibiting spatial statistics of different dispersion, in Fig.~\ref{fig:compare:regularity} (a). The artificial noises are added to the white background, the red rectangle on the left, the green rectangle on the middle, and the blue rectangle on the right with increasing degree of noises in order.
To preserve sharp boundaries, one has to manually choose a small regularization; however, large intensity variance in some of the data yields undesirably irregular boundaries between regions, with red and blue scattered throughout the middle and right rectangles (d), all of which however have sharp corners. On the other hand, to ensure homogeneity of the regions, one has to impose a large regularization, resulting in a biased final solution where corners are rounded (e), even for regions that would allow fine-scale boundary determination (red). Our approach with the adaptive regularization (b), however, naturally finds a solution with a sharp boundary where the data term supports it (red), and let the regularizer weigh-in on the solution when the data is more uncertain (blue). The zoom in images for the marked regions in (b) and (e) are shown in (c) and (f), respectively in order to highlight the geometric property of the solution around the corners.
\subsubsection{Multi-Label Segmentation on Real Images} \label{sec:experiment:segmentation:real}
%
%
\def\fw{68pt}
\def\sp{1pt}
\def\dexp{exp5}
\def\sa{62096}
\def\sal{4}
\def\sb{124084}
\def\sbl{3}
\def\sc{15062}
\def\scl{4}
\def\sd{238011}
\def\sdl{3}
\def\se{12003}
\def\sel{3}
\def\sf{29030}
\def\sfl{5}
\begin{figure*} [!th]
\centering
\includegraphics{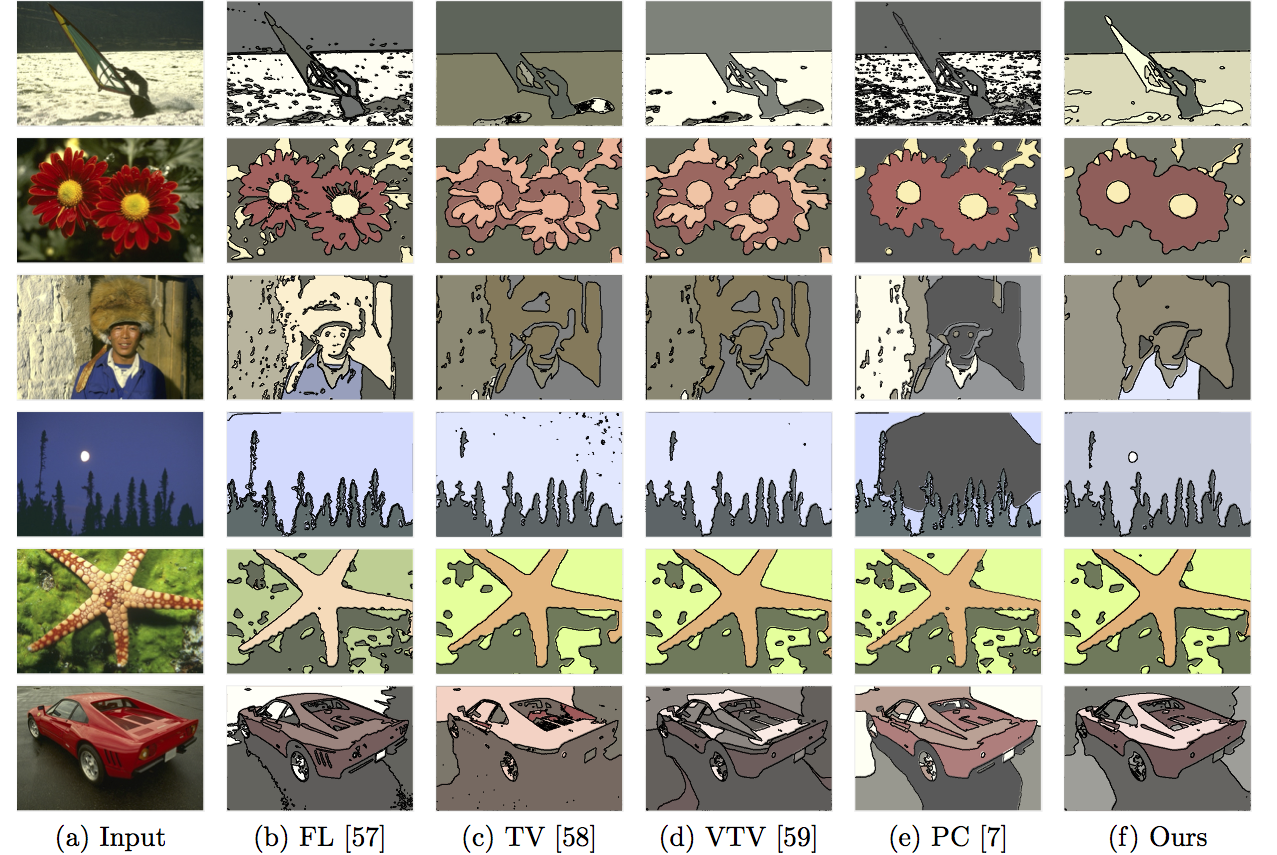}
\caption{Visual comparison of the multi-label segmentation on Berkeley dataset using different algorithms.}
\label{fig:berkeley}
\end{figure*}
%
%
%
\begin{table*}[htb]
\centering
{\footnotesize
\begin{tabular}{|c || c|c|c|c|c || c|c|c|c|c |}
\hline
\multirow{2}{*}{labels}& \multicolumn{5}{c||}{Precisioin} & \multicolumn{5}{c|}{Recall} \\\cline{2-11}
& FL~\cite{sundaramoorthi2014fast} & TV~\cite{zach2008fast} & VTV~\cite{lellmann:continuous:siam:2011} & PC~\cite{chambolle2012convex} & Ours & FL~\cite{sundaramoorthi2014fast} & TV~\cite{zach2008fast} & VTV~\cite{lellmann:continuous:siam:2011} & PC~\cite{chambolle2012convex} & Ours \\ \hline
3&0.53$\pm$0.11&0.68$\pm$0.17&0.68$\pm$0.16&0.57$\pm$0.15&0.67$\pm$0.13&0.78$\pm$0.11&0.67$\pm$0.10&0.67$\pm$0.11&0.76$\pm$0.11&0.69$\pm$0.12\\
4&0.48$\pm$0.08&0.53$\pm$0.18&0.58$\pm$0.31&0.57$\pm$0.20&0.63$\pm$0.134&0.84$\pm$0.08&0.71$\pm$0.04&0.75$\pm$0.08&0.76$\pm$0.14&0.72$\pm$0.08\\
5&0.44$\pm$0.12&0.43$\pm$0.30&0.50$\pm$0.22&0.49$\pm$0.16&0.60$\pm$0.05&0.89$\pm$0.06&0.81$\pm$0.07&0.74$\pm$0.11&0.71$\pm$0.18&0.73$\pm$0.13\\
6&0.42$\pm$0.10&0.37$\pm$0.24&0.43$\pm$0.16&0.47$\pm$0.14&0.52$\pm$0.19&0.82$\pm$0.09&0.71$\pm$0.15&0.71$\pm$0.10&0.72$\pm$0.10&0.65$\pm$0.22\\ \hline
\end{tabular}
}
\\\mbox{}\\
\caption{Precision and Recall of the multi-label segmentation results with varying number of labels.}
\label{tab:label}
\end{table*}
%
%
We compare our algorithm to the existing state-of-the-art techniques of which the underlying model assumes the piecewise constant image for fair comparison, and consider the algorithms: FL~\cite{sundaramoorthi2014fast}, TV~\cite{zach2008fast}, VTV~\cite{lellmann:continuous:siam:2011}, PC~\cite{chambolle2012convex}. 
%
We provide the qualitative evaluation in Fig.~\ref{fig:berkeley} where the input images are shown in (a) and the segmentation results are shown in (b)-(f) where the same number of labels is applied to all the algorithms.
The parameters for the algorithms under comparison are optimized with respect to the accuracy while we set the parameters for our algorithm: $\mu$=0.5, $\eta$=0.5, $\alpha$=0.01, $\beta$=10, $\tau$=0.5, $\theta$=1.
While our method yields better labels than the others, the obtained results may seem to be imperfect in general, which is due to the limitation of the underlying image model in particular in the presence of texture or illumination changes.
The quantitative comparisons are reported in terms of precision and recall with varying number of labels in Tables~\ref{tab:label}.
%
%
The computational cost as a baseline for $481 \times 321 \times 3$ color images without special hardware (e.g. multi-core GPU/CPU) and image processing techniques (e.g. image pyramid) is provided in Table~\ref{tab:cost}.
%
%
%
\begin{table}[htb]
\centering
\scriptsize
\begin{tabular}{|c|c|c|c|c|c|c|c|c|}
\hline
\# of labels&2&3&4&5&6&7&8&9\\
\hline
time (sec)&3.08&4.40&5.82&7.16&8.62&10.07&11.66&12.72\\
\hline
\end{tabular}
\\\mbox{}\\
\caption{Computational cost with varying number of labels.}
\label{tab:cost}
\end{table}
%
%
%
%
\subsection{Optical Flow} \label{sec:experiment:motion}
In the experiments, the qualitative and quantitative evaluation is performed based on the Middlebury optical flow dataset~\cite{baker2011database}. 
We use the average endpoint error (AEE)~\cite{otte1994optical} and the average angular error (AAE)~\cite{barron1994performance} for the quantitative evaluation.
%
\subsubsection{Effectiveness of Annealing in Warping} \label{sec:experiment:motion:anneal}
%
%
\def\fh{95pt}
\def\sp{3pt}
\begin{figure}[htb]
\centering
\begin{tabular}{c@{}c}
\includegraphics[totalheight=\fh]{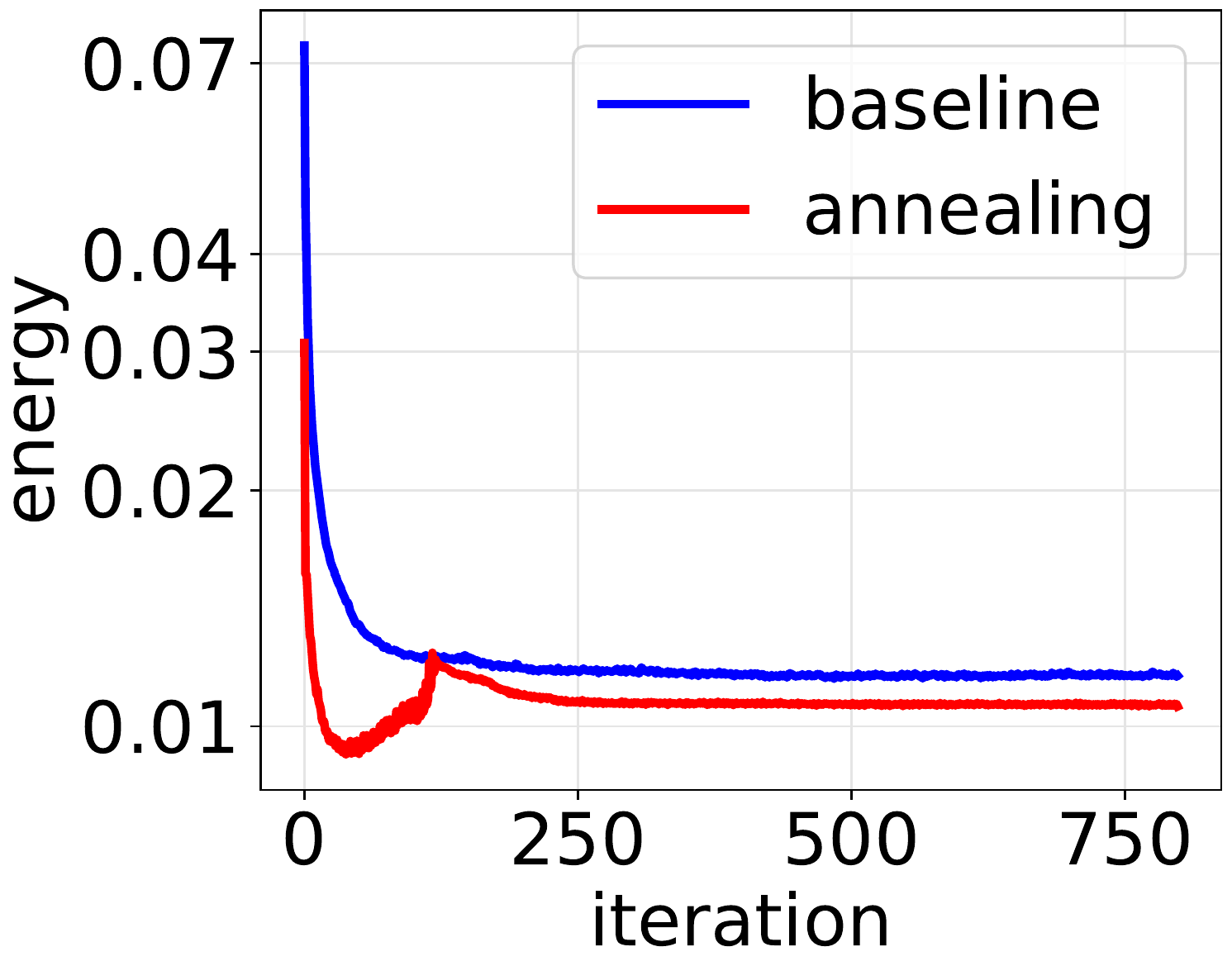} \hspace{\sp} &
\includegraphics[totalheight=\fh]{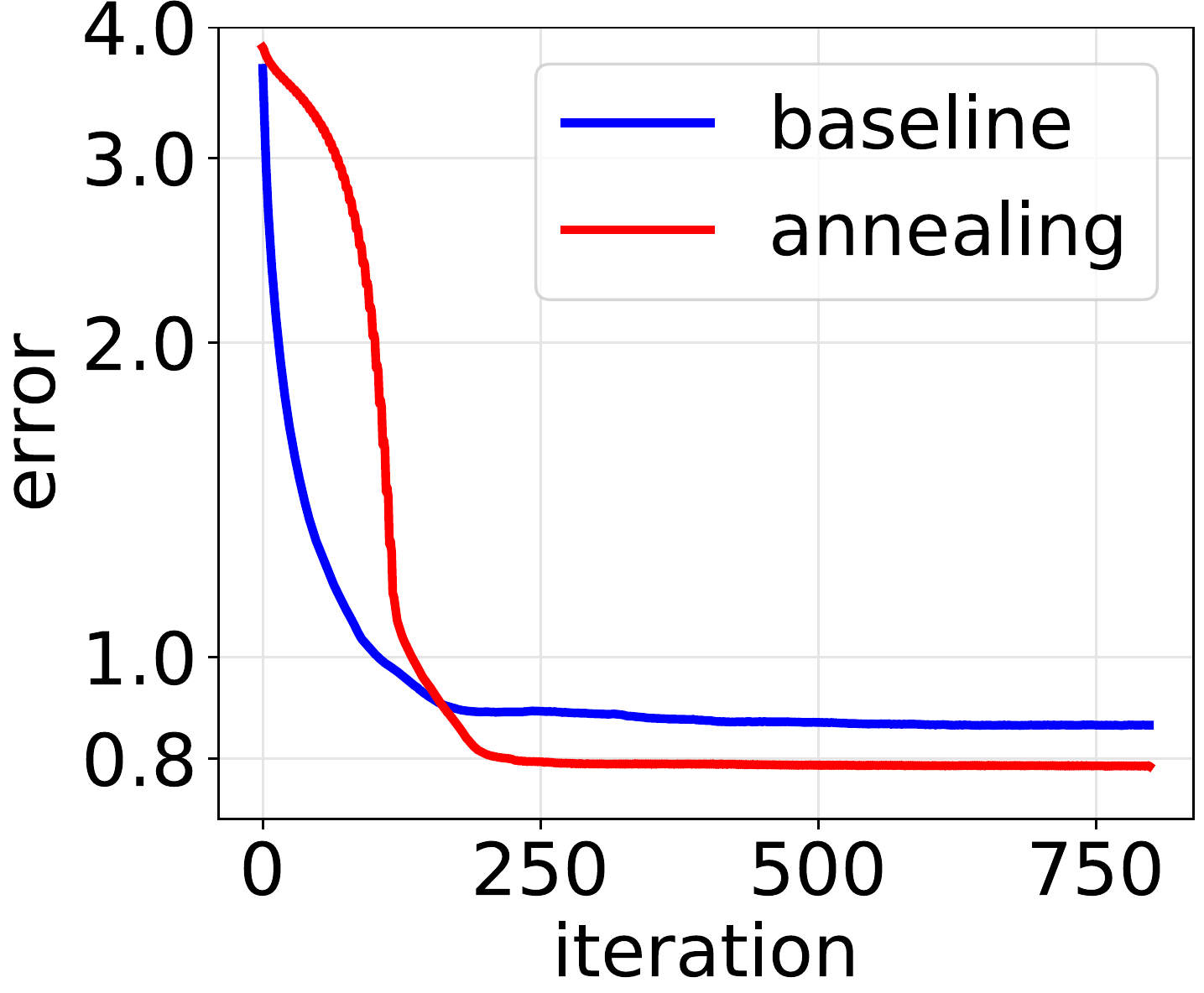}\\
(a) Energy & (b) End-point error
\end{tabular}
\caption{(a) Energy and (b) average end-point error comparison in motion estimation with and without annealing scheme.}
\label{fig:motion_annealing}
\end{figure}
We demonstrate the effectiveness of the annealing scheme for the degree of warping between the forward and the backward directions. 
We apply our optical flow algorithm without the use of the adaptive regularization in order to highlight the role of the annealing parameter, and use a pair of images with the largest disparity on average (Grove3 sequence) in the dataset. 
The quantitative comparison of the results is performed by our motion estimation algorithm with and without the use of the warping annealing parameter $\tau$ in Fig.~\ref{fig:motion_annealing} where (a) the energy and (b) the average endpoint error are presented. 
It is shown that the algorithm with the annealing of the warping yields faster convergence and better accuracy in comparison to the baseline without the warping annealing scheme. 
Note that the change of warping annealing parameter in optimization iteration results in the change of the energy, subsequently causing the fluctuation of the energy curve as shown in Fig.~\ref{fig:motion_annealing}(a) where $\Delta \tau = 0.005$ is used.
%
%
\def\fw{70pt}
\begin{figure*}[htb]
\centering
\includegraphics{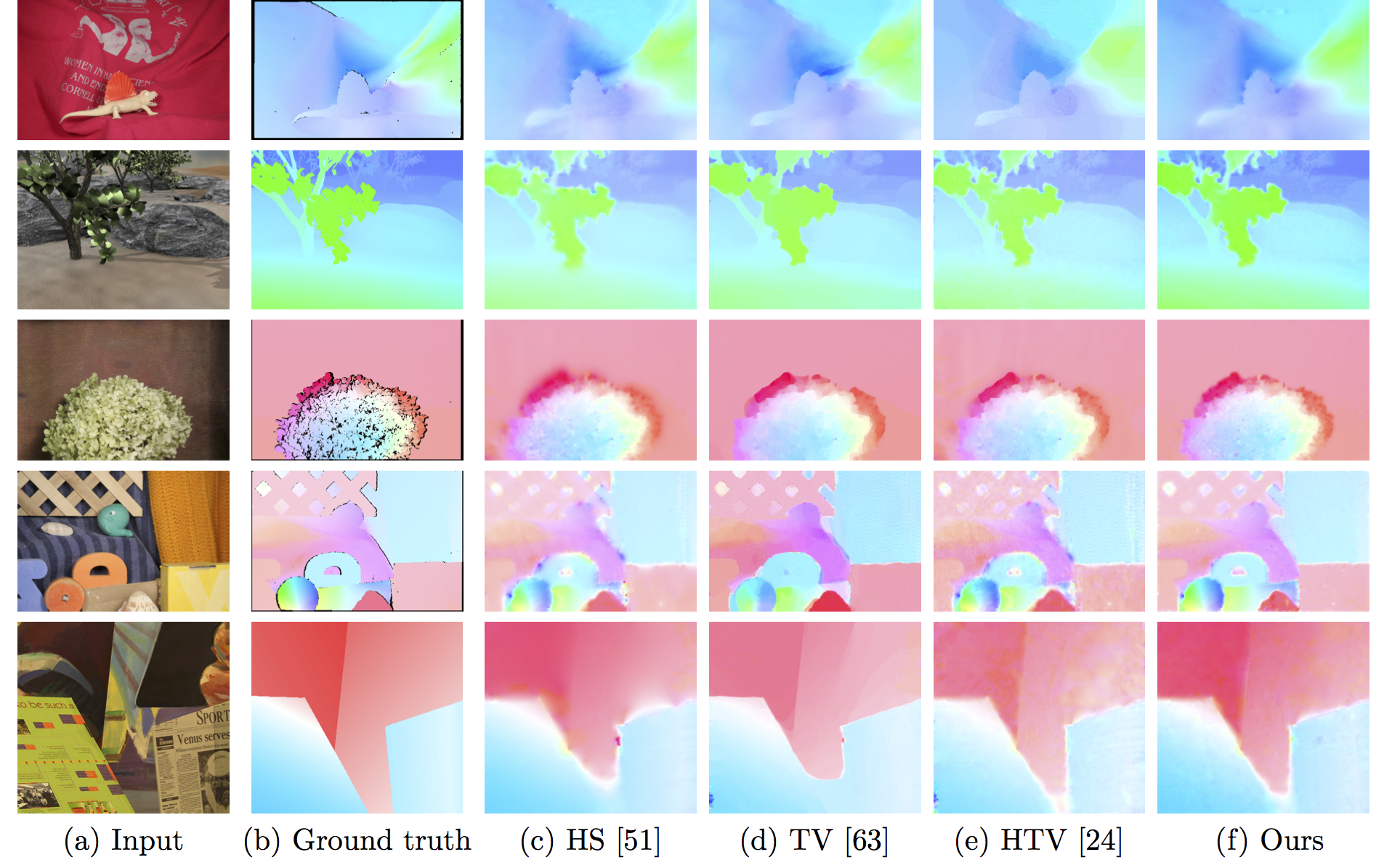}
\caption{Visual comparison of the motion estimation methods on Middlebury dataset using different algorithms.}
\label{fig:motion_visual}
\end{figure*}
\begin{table*}[htb]
\centering
{\scriptsize
\begin{tabular}{|c || c|c|c|c || c|c|c|c |}
\hline
\multirow{2}{*}{ Sequence }& \multicolumn{4}{c||}{Average End-point Error} & \multicolumn{4}{c|}{Average Angular Error} \\\cline{2-9}
& HS~\cite{horn1981determining} & TV~\cite{zach2007duality} & HTV~\cite{werlberger2009anisotropic} & Ours & HS~\cite{horn1981determining} & TV~\cite{zach2007duality} & HTV~\cite{werlberger2009anisotropic} & Ours \\ \hline
Dimetrodon 	& 0.1503 & 0.2345 & 0.1582 & 0.1270 & 0.0482 & 0.0754 & 0.0504 & 0.0433\\
Grove2 		& 0.2202 & 0.2292 & 0.2158 & 0.2279 & 0.0547 & 0.0574 & 0.0540 & 0.0569\\
Grove3 		& 0.8186 & 0.8267 & 0.7392 & 0.7494 & 0.1346 & 0.1411 & 0.1254 & 0.1323\\
Hydrangea 	& 0.3274 & 0.2521 & 0.2999 & 0.2027 & 0.0599 & 0.0528 & 0.0568 & 0.0414\\
RubberWhale & 0.2357 & 0.2566 & 0.2406 & 0.1468 & 0.1337 & 0.1388 & 0.1374 & 0.0806\\
Urban2 		& 0.7231 & 0.5867 & 0.5009 & 0.5098 & 0.0987 & 0.0745 & 0.0762 & 0.0819\\
Urban3 		& 1.1624 & 0.9345 & 0.9477 & 0.8771 & 0.1957 & 0.1470 & 0.1662 & 0.1436\\
Venus 		& 0.4066 & 0.4470 & 0.4175 & 0.4101 & 0.1185 & 0.1237 & 0.1168 & 0.1225\\ \hline
Average 	& 0.5055 & 0.4709 & 0.4399 & 0.4063 & 0.1055 & 0.1013 & 0.0979 & 0.0878\\ \hline
\end{tabular}
}
\\\mbox{}\\
\caption{The average end-point error and average angular error of the motion estimation results.}
\label{tab:motion_average}
\end{table*}
%
%
\subsubsection{Comparison to Classical Algorithms} \label{sec:experiment:motion:compare}
We compare our algorithm to the classical Horn-Schunck model (HS)~\cite{horn1981determining}, TV-$L_1$ model (TV) ~\cite{zach2007duality} and Huber variant of total variation with $L_1$ data fidelity (HTV)~\cite{werlberger2009anisotropic}, and these algorithms are optimized by primal-dual algorithm~\cite{chambolle2011first}.
The visualization for the computed velocity fields using the standard color coding scheme~\cite{baker2011database} are presented in Fig.~\ref{fig:motion_visual} where (a) the input images, (b) the ground truth, (c) HS, (d) TV, (e) HTV and (f) our method are shown.
These visual comparisons indicate that our algorithm is more precise than the others, which is quantitatively evaluated in Table~\ref{tab:motion_average} where AEE and AAE are computed for each case. 
Note that the occlusions are not explicitly taken into special consideration in the computation of the optical flow in order to emphasize the role of the adaptive regularization that implicitly deals with occlusions where higher residuals due to the mismatch occur. 
The parameters for each method are optimally selected with respect to the errors, and we use $\mu = 0.01$, $\eta = 0.3$, $\alpha = 0.01$, $\beta = 10$, $\Delta \tau = 0.005$, $\theta = 0.1$ for our algorithm.
%
%
\subsection{Denoising} \label{sec:experiment:denoise}
\def\fH{110pt}
\def\case{124084}
\begin{figure*}[htb]
\centering
\begin{tabular}{c@{ }c@{ }c}
\includegraphics[height=\fH]{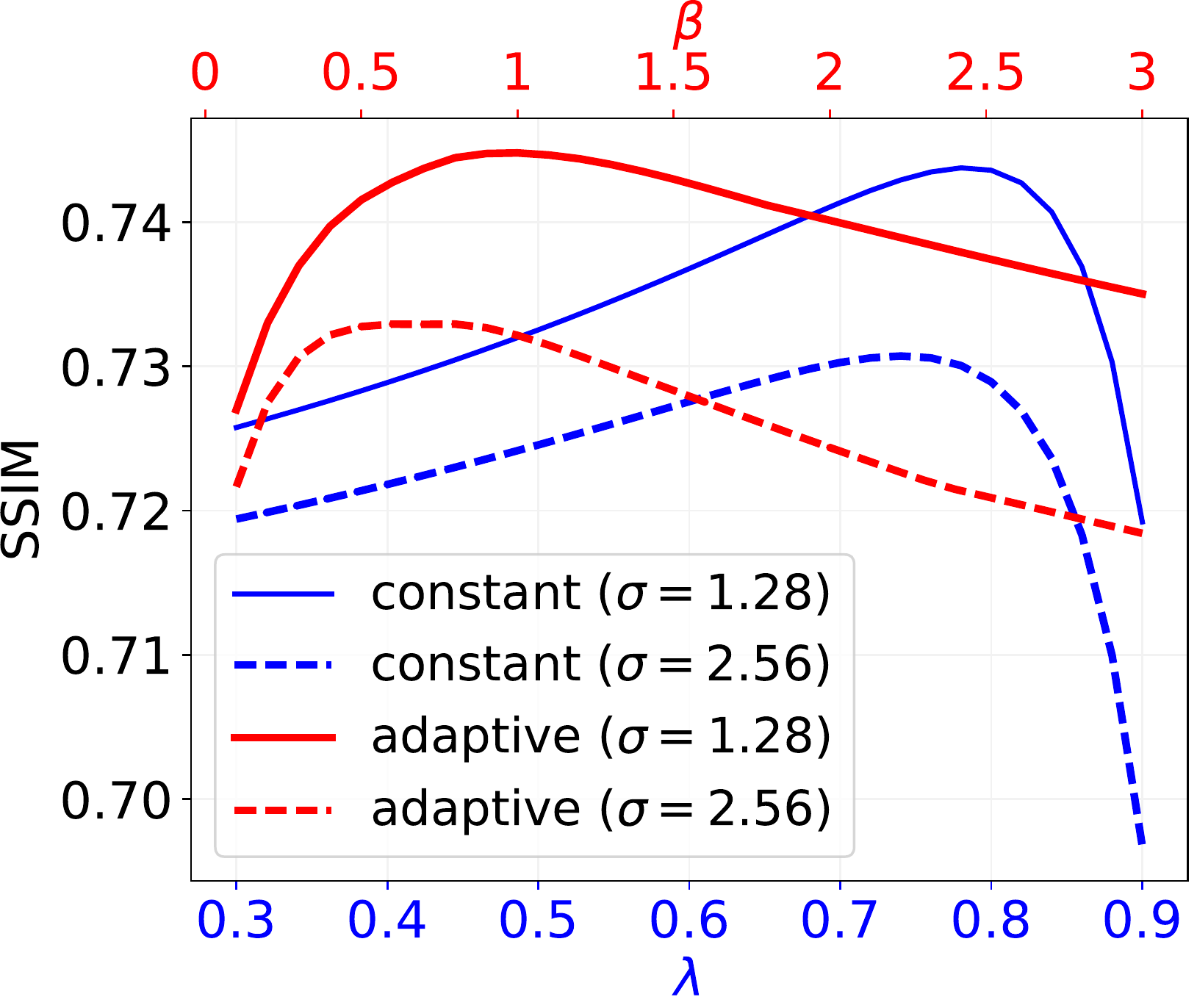} \hspace{0pt} & \hspace{0pt}%
\includegraphics[height=\fH]{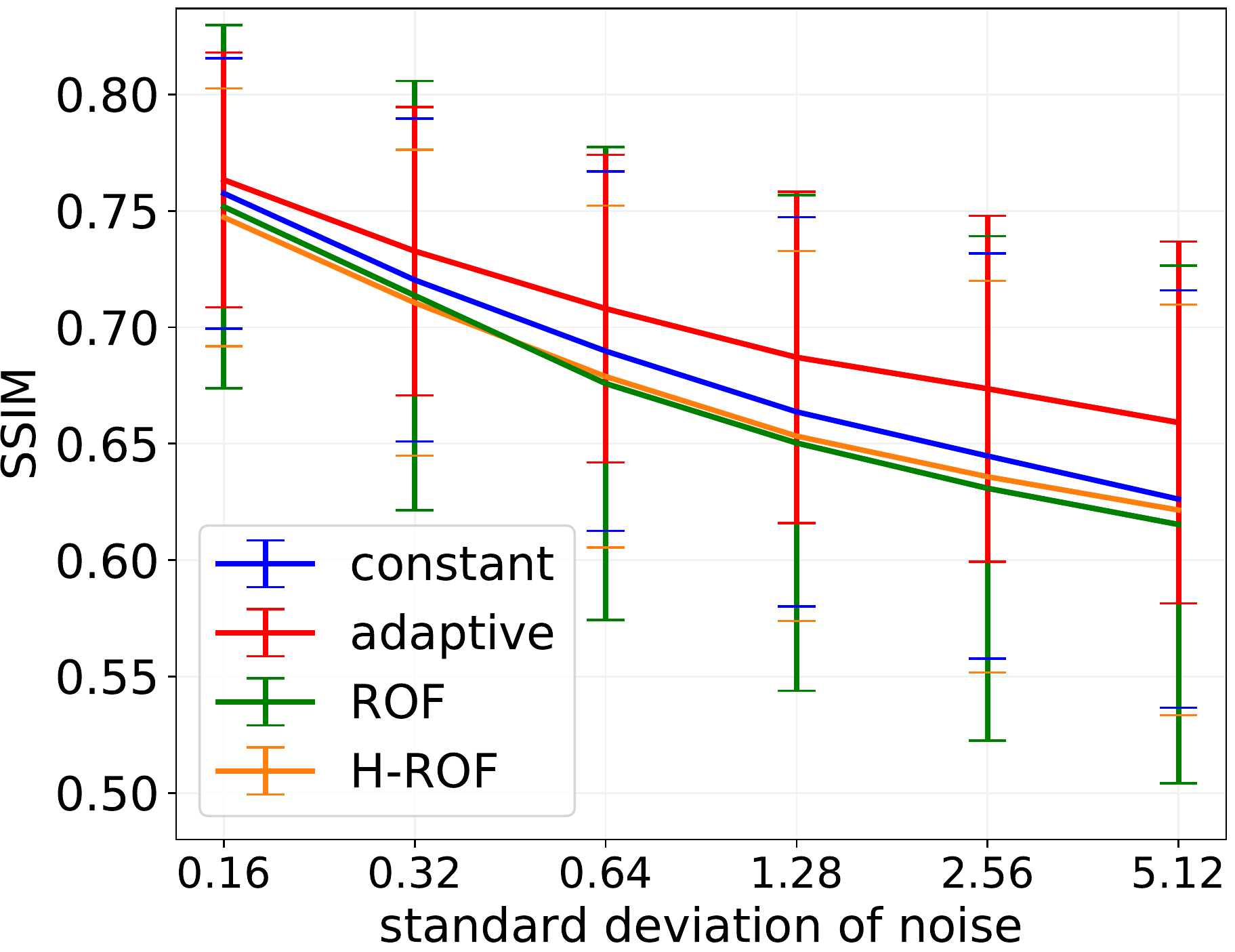} \hspace{0pt} & \hspace{0pt} %
\includegraphics[height=\fH]{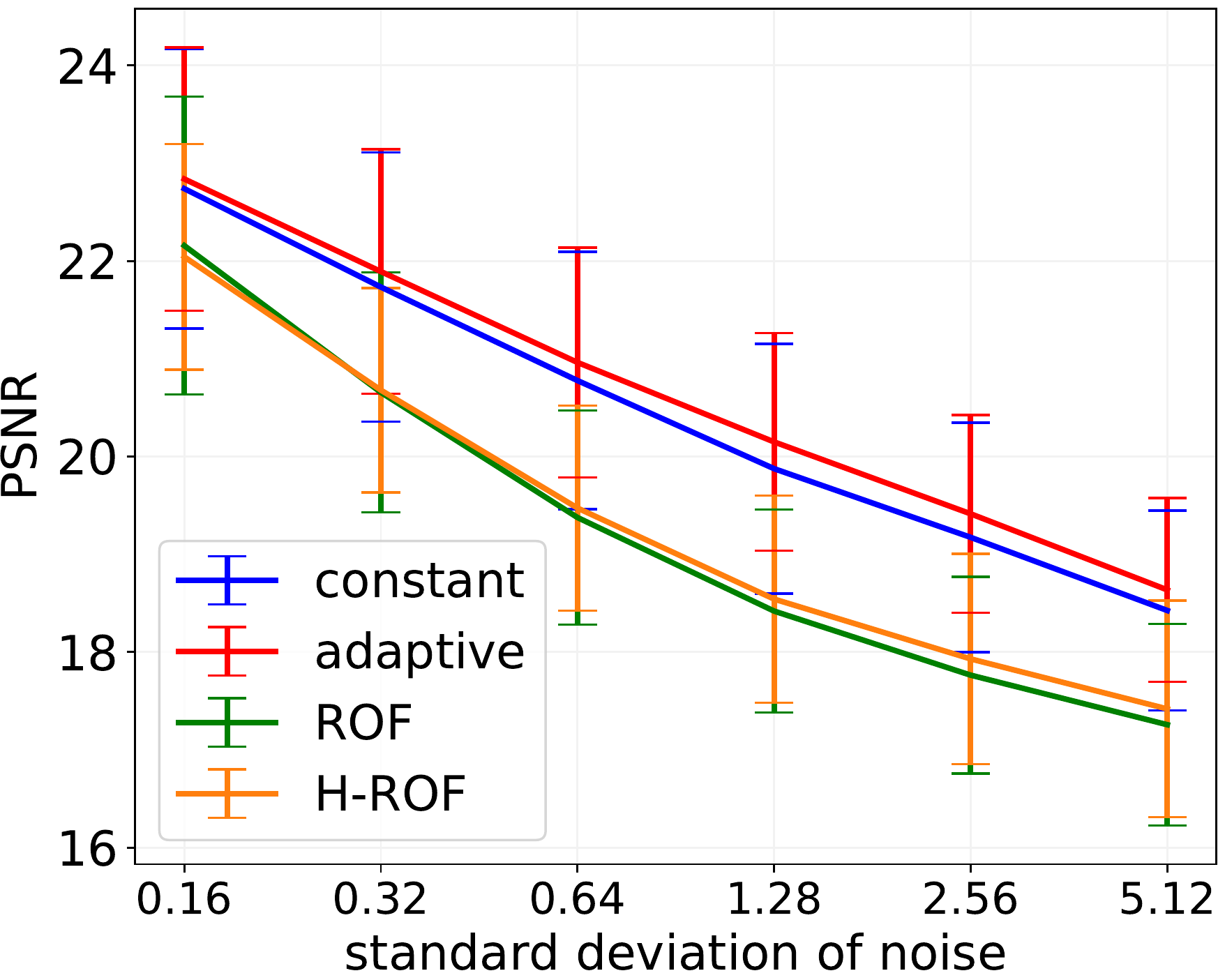}\\[-\dp\strutbox] 
(a) SSIM v.s. ($\lambda, \beta$) \hspace{0pt} & \hspace{0pt} (b) SSIM v.s. noise $\sigma$ \hspace{0pt} & \hspace{0pt} (c) PSNR v.s. noise $\sigma$
\end{tabular}
\caption{(a) Error measure with varying regularization parameters for SSIM, where the bottom x-axis represents $\lambda$ (constant) and the top x-axis represents $\beta$ (adaptive). The comparative results with respect to (b) PSNR and (c) SSIM for the images with varying noise standard deviations (x-axis).}
\label{fig:denoise_param}
\end{figure*}
\def\seqa{boat.512}
\def\seqb{4.1.06}
\def\seqc{4.2.07}
\def\seqd{4.1.05}
\def\stda{0.64}
\def\stdb{0.64}
\def\stdc{0.64}
\def\stdd{0.64}
\def\fw{70pt}
\def\sp{5pt}
\begin{figure*}[htb]
\centering
\includegraphics{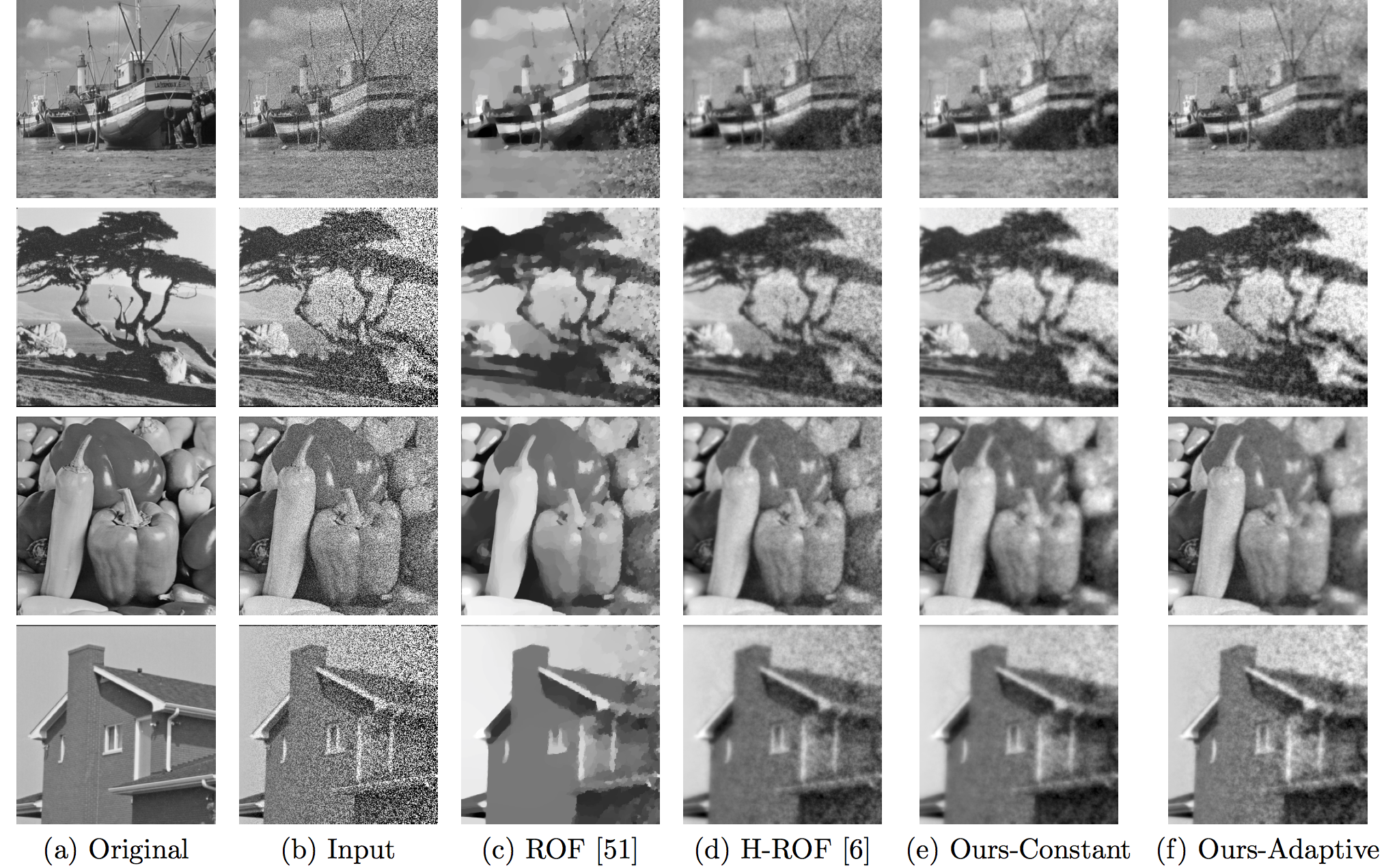}
\caption{Visual comparison of the denoising results using different algorithms for the input images (b) having spatially biased Gaussian noises from the original images (a).}
\label{fig:denoise_image}
\end{figure*}
\begin{table*}[htb]
\centering
{\scriptsize
\begin{tabular}{|c || c|c|c|c || c|c|c|c |}
\hline
\multirow{2}{*}{Noise $\sigma$} & \multicolumn{4}{c||}{SSIM} & \multicolumn{4}{c|}{PSNR} \\\cline{2-9}
& ROF~\cite{rudin1992nonlinear} & H-ROF~\cite{chambolle2011first} & Ours-Constant & Ours-Adaptive & ROF~\cite{rudin1992nonlinear} & H-ROF~\cite{chambolle2011first} & Ours-Constant & Ours-Adaptive \\  \hline
0.16 & 0.7517$\pm$0.0780 & 0.7472$\pm$0.0554 & 0.7575$\pm$0.0580 & 0.7633$\pm$0.0547 & 22.1570$\pm$1.5226 & 22.0422$\pm$1.1537 & 22.7400$\pm$1.4307 & 22.8386$\pm$1.3464 \\
0.32 & 0.7136$\pm$0.0922 & 0.7105$\pm$0.0657 & 0.7203$\pm$0.0694 & 0.7327$\pm$0.0619 & 20.6562$\pm$1.2274 & 20.6787$\pm$1.0458 & 21.7331$\pm$1.3758 & 21.8920$\pm$1.2506 \\
0.64 & 0.6758$\pm$0.1016 & 0.6788$\pm$0.0734 & 0.6898$\pm$0.0772 & 0.7080$\pm$0.0661 & 19.3762$\pm$1.0956 & 19.4729$\pm$1.0496 & 20.7769$\pm$1.3158 & 20.9623$\pm$1.1757 \\
1.28 & 0.6503$\pm$0.1064 & 0.6533$\pm$0.0794 & 0.6637$\pm$0.0836 & 0.6871$\pm$0.0712 & 18.4191$\pm$1.0374 & 18.5419$\pm$1.0601 & 19.8757$\pm$1.2771 & 20.1495$\pm$1.1142 \\
2.56 & 0.6308$\pm$0.1083 & 0.6358$\pm$0.0841 & 0.6447$\pm$0.0870 & 0.6736$\pm$0.0743 & 17.7635$\pm$1.0052 & 17.9296$\pm$1.0747 & 19.1732$\pm$1.1733 & 19.4138$\pm$1.0119 \\
5.12 & 0.6153$\pm$0.1111 & 0.6215$\pm$0.0882 & 0.6262$\pm$0.0896 & 0.6591$\pm$0.0777 & 17.2568$\pm$1.0293 & 17.4183$\pm$1.1071 & 18.4251$\pm$1.0205 & 18.6358$\pm$0.9411 \\ \hline
\end{tabular}
}
\caption{The SSIM and PSNR of the denoising results with varying noise standard deviation $\sigma$.}
\label{tab:denoise}
\end{table*}
In the experiments, we compare our Huber-Huber model with constant and adaptive regularization to TV-$L_2$ model (ROF)~\cite{rudin1992nonlinear} and Huber-$L_2$ model (H-ROF)~\cite{chambolle2011first} in terms of the structural similarity (SSIM)~\cite{wang2004image} and the peak signal-to-noise ratio (PSNR).
We apply the denoising algorithm to the images in the USC-SIPI dataset~\cite{weber1997usc} with spatially biased Gaussian noise of different noise levels.
We first demonstrate the effectiveness of our adaptive regularization in Fig.~\ref{fig:denoise_param} (a) where SSIM is computed with varying regularization parameters ($\beta$ for the adaptive in red at top axis, and $\lambda$ for the constant in blue at bottom axis) for the images with different noise levels (noise $\sigma$ is 1.28 in solid line, and 2.56 in dotted line).
In Fig.~\ref{fig:denoise_param}, we quantitatively evaluate our $\mathrm H^2$ model with constant and adaptive regularization in comparison to TV-$L_2$ model (ROF)~\cite{rudin1992nonlinear} and Huber-$L_2$ model (H-ROF)~\cite{chambolle2011first} in terms of (b) SSIM and (c) PSNR for the images with different degree of spatially varying noises.
It is clearly shown that our method with adaptive regularization yields better SSIM and PSNR than the other methods over all the noise levels.
The visual comparison of the results is provided in Fig.~\ref{fig:denoise_image} where (a) original images are shown, (b) input noisy images, (c) results by TV-$L_2$ model~\cite{rudin1992nonlinear}, (d) Huber-$L_2$ model~\cite{chambolle2011first}, (e) our $\mathrm H^2$ with constant regularization, and (f) our $\mathrm H^2$ with adaptive regularization, where the results are optimized with respect to SSIM and the results are similar to PSNR.
The results by conventional models in (c), (d), (e) indicate that undesired excessive smoothing is globally applied to cope with the highest noise level that is locally present.
In contrast, our algorithm with adaptive regularization yields the results where the degree of smoothing is adaptively determined by the spatially varying local residuals.
Note that the presented visual results in Fig.~\ref{fig:denoise_image} may not seem perfect since images with relatively high noises are used to distinguish the algorithm characteristics.
The quantitative comparison of the algorithms with varying degrees of spatially varying noises is provided in Table~\ref{tab:denoise} where the parameters for each algorithm are carefully chosen to yield the best performance for each error measure and we use $\mu = 0.16$, $\eta = 0.08$, $\alpha = 0.01$, $\beta = 1$, $\theta = 1$ for our algorithm.
%
%
%
\section{Conclusion} \label{sec:conclusion}
%
We have introduced a novel regularization algorithm in a variational framework where a composite energy functional is optimized.
Our scheme weighs a prior, or regularization functional, depending both on time (iteration) during the convergence procedure, and on the local spatial statistics of the data. This results in a natural annealing schedule whereby the influence of the prior is strongest at initialization, and wanes as the solution approaches a good fit with the data term. It imposes regularization where needed, and lets the data drive the process when it is sufficiently informative. 

All this is done within an efficient convex optimization framework using ADMM.
We have proposed an energy functional that uses the Huber function as a robust loss estimator for both data fidelity and regularization.
An efficient optimization algorithm has been applied with a variable splitting, which has yielded faster and more accurate solutions in comparison to the conventional models based on Total Variation.
The adaptive regularization has been demonstrated to be more effective for the classical imaging problems including segmentation, motion estimation and denoising in particular when the distribution of degrading factors is spatially biased. 
%
%
\appendices
%
%
\section{Proof} \label{sec:proof}
\subsection{Proof of Theorem~\ref{thm:unique}} \label{sec:proof:theorem}

We provide a sketch of the fixed point argument in the following.  The topology we use is strong convergence of $(u,\rho(u))$ in $L^1(\Omega)\times L^1(\Omega)$, and we construct a self-map on this space. Then the map $\u \in L^1(\Omega) \mapsto \rho(u) \in L^1(\Omega)$ is trivially continuous.
With the properties of the convolution kernel $G$ we immediately see that the map $\rho \in L^1(\Omega) \mapsto G * \rho \in C^1(\Omega)$ is continuous and compact. Moreover, the map $ G * \rho \in C^1(\Omega) \mapsto \lambda \in C^1(\Omega)$ is continuous. Finally we see from a standard continuous dependence argument on the variational problem that $\lambda \in C^1(\Omega) \mapsto (u,\rho(u)) \in
BV_0(\Omega) \times L^1(\Omega)$ is continuous, and the 
 continuous embedding of $BV_0(\Omega)$ into $L^1(\Omega)$ finally implies the continuity and compactness and fixed point operator on these spaces. In order to apply a Schauder's fixed-point theorem and conclude the existence of a fixed point, it suffices to show that some bounded set is mapped into itself. For this sake let $C_0 = \int_\Omega \rho(0) \ud x$, and choose $c$ such that
$$ c \leq \exp\left(-\frac{\Vert G\Vert_\infty C_0}{\beta c}\right). $$
The existence of such a constant $c$ is guaranteed for $\beta$ sufficiently large. Now let 
$$ \int \rho(u) \ud x \leq \frac{C_0}c, $$ then we obtain with a standard estimate of the convolution and monotonicity of the exponential function that
$$ c \leq \lambda = \exp\left(-\frac{G*\rho}\beta\right) \leq 1. $$
Moreover, there exists a constant $\tilde c$ such that $\tilde c \leq 1-\lambda \leq 1$. Hence, a minimizer $u$ of $E_\lambda$ satisfies
$$ c \int_\Omega \rho(u) \ud x + \tilde c \int_\Omega |\nabla u| \ud x \leq E_\lambda(u) \leq E_\lambda(0) \leq \int_\Omega \rho(0) \ud x,$$
where $\int_\Omega \rho(0) \ud x = C_0$.
Hence using the closed set of $u,\rho$ such that 
$$ \Vert \rho \Vert_{L^1} \leq \frac{C_0}{c}, \qquad \vert u \vert_{BV_0} \leq \frac{C_0}{\tilde c}, $$
we obtain a self-mapping by our fixed-point operator.
%
%
%
\section{Optimization Algorithm} \label{sec:optimality}
%

\subsection{Optimality Conditions for Segmentation} \label{sec:optimality:segmentation}
The optimization steps for minimizing the Lagrangian~\eqref{eq:EnergyUnconstrained} are summarized in Algorithm~\ref{alg:admm:segmentation}.
%
%
The update of the estimate $c_i^{k+1}$ in~\eqref{eq:EnergyUnconstrainedLabel} is obtained by:
\begin{align} 
\quad c_i^{k+1} \coloneqq \frac{\int_\Omega \lambda_i^k \, (f - r_i^k) \, u_i^k \ud x}{\int_\Omega \lambda_i^k \, u_i^k \ud x}. \label{step:SolutionC:appen}
\end{align}
The update for the auxiliary variable $r_i^{k+1}$ in~\eqref{eq:DataFidelityMoreauYosida} is obtained by:
\begin{align} 
0 & \in \partial | r_i^{k+1} | - \frac{1}{\mu} ( f - c_i^{k+1} - r_i^{k+1} ), \label{eq:OptimalityR}
\end{align}
where $\partial$ denotes the sub-differential operator. 
The solution for the optimality condition in~\eqref{eq:OptimalityR} is obtained by the soft shrinkage operator: 
%
\begin{align}
r_i^{k+1} \coloneqq \mathcal{T} \left( f - c_i^{k+1} \, \big| \, \mu \right). \label{eq:shrink_r}
\end{align}
Similarly, the update of the weighting function $\lambda_i^{k+1}$ in~\eqref{eq:EnergyUnconstrainedLabel} and the auxiliary variable $z_i^{k+1}$ in~\eqref{eq:RegularizationMoreauYosida} are obtained by:
\begin{align}
\lambda_i^{k+1} & \coloneqq \mathcal{T} \left( \nu_i^{k+1} \, \big| \, \alpha \right), \label{eq:shrink_l}\\
z_i^{k+1} & \coloneqq \mathcal{T} \left( \nabla v_i^k \, \big| \, \eta \right), \label{eq:shrink_z}
\end{align}
where $\nu_i^{k+1}$ is computed in~\eqref{eq:composite:weight:nu}.
For the update of the primal variable $u_i^{k+1}$, we employ the intermediate solution $\tilde{u}_i^{k+1}$ of which the optimality condition is given by:
\begin{align} 
0 \in \lambda_i^{k+1} d_i^{k+1} + \tau \sum_{i \neq j} u_j^k + \theta (\tilde{u}_i^{k+1} - v_i^k + w_i^k), \label{eq:StepUCondition:appen}\\
d_i^{k+1} \coloneqq | r_i^{k+1} | + \frac{1}{2 \mu} (f - c_i^{k+1} - r_i^{k+1})^2, \label{eq:StepUConditionD}
\end{align}
leading to the following update:
\begin{align} 
\tilde{u}_i^{k+1} \coloneqq v_i^k - w_i^k - \frac{\lambda_i^{k+1}}{\theta} d_i^{k+1} - \frac{\tau}{\theta} \sum_{i \neq j} u_j^k. \label{eq:StepUInter}
\end{align}
Given the intermediate solution $\tilde{u}_i^{k+1}$, the positivity constraint is imposed for the update of $u_i^{k+1}$:
\begin{align} 
u_i^{k+1} \coloneqq \Pi_A( \tilde{u}_i^{k+1} ) = \max \{ 0, \tilde{u}_i^{k+1} \}, \label{eq:StepU}
\end{align}
where the orthogonal projection operator $\Pi_A$ on a set $A = \{ x \, | \, x \ge 0 \}$ is defined by:
\begin{align} \label{eq:projection:appen}
\Pi_A(x) = \arg\min_{y \in A} \| y - x \|_2.
\end{align}
We also employ the intermediate solution $\tilde{v}_i^{k+1}$ for the update of the primal variable $v_i^{k+1}$ and its optimality condition reads:
\begin{align} 
0 \in (1 - \lambda_i^{k+1}) \nabla^* ( \nabla \tilde{v}_i^{k+1} - z_i^{k+1} ) - \eta \theta( u_i^{k+1} - \tilde{v}_i^{k+1} + w_i^k ), \nonumber
\end{align}
where $\nabla^*$ denotes the adjoint operator of $\nabla$, leading to the following linear system of equation with $\xi_i = \frac{1 - \lambda_i^{k+1}}{\eta \theta}$:
\begin{align} 
\tilde{v}_i^{k+1} -  \xi_i \Delta \tilde{v}_i^{k+1} \coloneqq u_i^{k+1} + w_i^{k} - \xi_i \, \dv{z_i^{k+1}}, \label{eq:StepV}
\end{align}
where $- \nabla^* \nabla = \Delta$ is the Laplacian operator, and $- \nabla^* = \mbox{div}$ is the divergence operator. 
We use the Gauss-Seidel iterations to solve this linear system of equation.
Given the set of intermediate solution $\{ \tilde{v}_i^{k+1} \}$, the solution for the update of the variable $v_i^{k+1}$ is obtained by the orthogonal projection of the intermediate solution to the set $B$:
\begin{align} 
v_i^{k+1} \coloneqq \tilde{v}_i^{k+1} - \frac{1}{n} \left( V - 1 \right), \quad
V = \sum_{i \in \Lambda} \tilde{v}_i^{k+1}.
\end{align}
The update of the dual variable $w_i^{k+1}$ is obtained by the gradient ascent scheme:
\begin{align} 
w_i^{k+1} = w_i^k + u_i^k - v_i^k. \label{eq:stepW}
\end{align}
%

%
\begin{algorithm}[tb]
\caption{The ADMM updates for minimizing~\eqref{eq:EnergyUnconstrained}}
\label{alg:admm:segmentation}
\begin{algorithmic}
\State 
\textbf{for} each label $i \in \Lambda$ \textbf{ do}
\begin{flalign}
\nu_i^{k+1} & \coloneqq \exp\left( - \frac{\rho(u_i^k, c_i^k, r_i^k)}{\beta} \right) & \label{step:nu}\\
\lambda_i^{k+1} & \coloneqq \argmin_{\lambda} \frac{1}{2} \| \nu_i^{k+1} - \lambda \|_2^2 + \alpha \| \lambda \|_1 & \label{step:l}\\
\quad c_i^{k+1} & \coloneqq \argmin_{c} \rho( u_i^k, c, r_i^k ) & \label{step:c}\\
r_i^{k+1} & \coloneqq \argmin_{r} \rho( u_i^k, c_i^{k+1}, r ) & \label{step:r}\\
z_i^{k+1} & \coloneqq \argmin_z \gamma(v_i^k, z) & \label{step:z}\\
u_i^{k+1} & \coloneqq \argmin_{u} \int_\Omega \lambda_i^{k+1} \rho(u, c_i^{k+1}, r_i^{k+1}) \ud x + \delta_A(u) & \nonumber\\
& + \int_\Omega \tau \Big( \sum_{i \neq j} u_j \Big) u \ud x + \frac{\theta}{2} \| u - v_i^k + w_i^k \|_2^2  & \label{step:u}\\
\tilde{v}_i^{k+1} & \coloneqq \argmin_{v} \int_\Omega \left( 1-\lambda_i^{k+1} \right) \gamma(v, z_i^{k+1}) \ud x & \nonumber\\
& + \frac{\theta}{2} \| u_i^{k+1} - v + w_i^k \|_2^2 & \label{step:v}\\
w_i^{k+1} & \coloneqq w_i^k + u_i^{k+1} - v_i^{k+1} & \label{step:w}
\end{flalign}
\textbf{end for}
\begin{flalign}
\{ v_i^{k+1} \} & \coloneqq \Pi_B \left( \{ \tilde{v}_i^{k+1} \} \right) & \label{step:v:projection}
\end{flalign}
\end{algorithmic}
\end{algorithm}
%
%

\subsection{Optimality Conditions for Optical Flow} \label{sec:optimality:motion}
%
%
\begin{algorithm}[tb]
\caption{The ADMM updates for minimizing~\eqref{eq:energy:motion:lagrangian}}
\label{alg:admm:motion}
\begin{algorithmic}
\State 
\begin{flalign}
	\nu^{k+1} & \coloneqq \exp\left( - \frac{\rho_\tau(u^k, r^k)}{\beta} \right) & \label{motion:step:nu}\\
	\lambda^{k+1} & \coloneqq \argmin_{\lambda} \frac{1}{2} \| \nu^{k+1} - \lambda \|_2^2 + \alpha \| \lambda \|_1 & \label{motion:step:l}\\
	r^{k+1} & \coloneqq \argmin_{r} \rho_\tau( u^k, r ) & \label{motion:step:r}\\
\end{flalign}
\textbf{for} \textrm{ each component } $i \in \{1, 2\}$ \textbf{ do}
\begin{flalign}
	\quad z_i^{k+1} & \coloneqq \argmin_z \gamma(v_i^k, z) & \label{motion:step:z}\\
	u_i^{k+1} & \coloneqq \argmin_{u} \int_\Omega \lambda^{k+1} \, \rho_\tau(u, r^{k+1}) \dx & \nonumber\\
		& + \frac{\theta}{2} \| u - v_i^{k+1} + w_i^k \|_2^2  & \label{motion:step:u}\\
	v_i^{k+1} & \coloneqq \argmin_{v} \int_\Omega (1 - \lambda^{k+1}) \, \gamma(v, z_i^{k+1}) \dx & \nonumber\\
		& + \frac{\theta}{2} \| u_i^k - v + w_i^k \|_2^2  & \label{motion:step:v}\\
	w_i^{k+1} & \coloneqq w_i^k + u_i^{k+1} - v_i^{k+1} & \label{motion:step:w}
\end{flalign}
	\textbf{end for}
\begin{flalign}
	\tau^{k+1} & \coloneqq \min(1, \tau^k + \Delta \tau) & \label{motion:step:tau} 
\end{flalign}
\end{algorithmic}
\end{algorithm}
%
%
%
%
%
The ADMM update steps for minimizing~\eqref{eq:energy:motion:lagrangian} are summarized in Algorithm~\ref{alg:admm:motion}.
The update for the variables $r^{k+1}$ in~\eqref{eq:motion:data:warp:rho:MoreauYosida} is obtained by the soft shrinkage operator:
\begin{align}
r^{k+1} & \coloneqq \mathcal{T} \left( f_t - ( \nabla f_1 + \tau^k \nabla f_2 ) u^k \, \big| \, \mu \right). \label{eq:motion:shrink_r}
\end{align}
The variables $\lambda^{k+1}$ in~\eqref{eq:energy:motion:split} and $z_i^{k+1}$ in~\eqref{eq:motion:gamma:MoreauYosida} are updated in the same way as in~\eqref{eq:shrink_l} and~\eqref{eq:shrink_z}.
%
The optimality condition for the update of the primal variable $u_i^{k+1}$ reads:
\begin{align} 
&0 \in - \lambda^{k+1} d^{k+1} \, (\nabla f_1 + \tau^k \nabla f_2) + \mu \theta (u - v^k + w^k), \label{eq:motion:condition:u}\\
&d^{k+1} \coloneqq f_t - (\nabla f_1 + \tau^k \nabla f_2) u - r^{k+1}, \label{eq:motion:condition:d}
\end{align}
leading to the following solution:
\begin{align} 
(I + A A^T) u^{k+1} = \frac{\mu \theta}{\lambda^{k+1}} (v^k - w^k) + (f_t - r^{k+1}) A, \label{eq:motion:stepU}
\end{align}
where $A = \nabla f_1 + \tau^k \nabla f_2$ and $I$ denotes the identity matrix.
The solutions for the update of $v_i^{k+1}$ and $w_i^{k+1}$ are obtained in the same way as in~\eqref{eq:StepV} and~\eqref{eq:stepW}, respectively.
%
%

%
%
\subsection{Optimality Conditions for Denoising} \label{sec:optimality:denoise}
%
%
\begin{algorithm}[tb]
\caption{The ADMM updates for minimizing~\eqref{eq:energy:denoise:lagrangian}}
\label{alg:admm:denoise}
\begin{algorithmic}
\State 
\begin{flalign}
	\nu^{k+1} & \coloneqq \exp\left( - \frac{\rho(u^k, r^k)}{\beta} \right) & \label{denoise:step:nu}\\
	\lambda^{k+1} & \coloneqq \argmin_{\lambda} \frac{1}{2} \| \nu^{k+1} - \lambda \|_2^2 + \alpha \| \lambda \|_1 & \label{denoise:step:l}\\
	r^{k+1} & \coloneqq \argmin_{r} \rho( u^k, r ) & \label{denoise:step:r}\\
	z^{k+1} & \coloneqq \argmin_z \gamma(v^k, z) & \label{denoise:step:z}\\
	u^{k+1} & \coloneqq \argmin_{u} \int_\Omega \lambda^{k+1} \, \rho(u, r^{k+1}) \dx & \nonumber\\
		& + \frac{\theta}{2} \| u - v^{k+1} + w^k \|_2^2  & \label{denoise:step:u}\\
	v^{k+1} & \coloneqq \argmin_{v} \int_\Omega (1 - \lambda^{k+1}) \, \gamma(v, z^{k+1}) \dx & \nonumber\\
		& + \frac{\theta}{2} \| u^k - v + w^k \|_2^2  & \label{denoise:step:v}\\
	w^{k+1} & \coloneqq w^{k} + u^{k+1} - v^{k+1} & \label{denoise:step:w}
\end{flalign}
\end{algorithmic}
\end{algorithm}
%
%
%
The ADMM update steps for minimizing~\eqref{eq:energy:denoise:lagrangian} are summarized in Algorithm~\ref{alg:admm:denoise}.
The update for the variables $r^{k+1}$ in~\eqref{eq:denoise:data:MoreauYosida} is obtained by the soft shrinkage operator:
\begin{align}
r^{k+1} & \coloneqq \mathcal{T} \left( f - u^k \, \big| \, \mu \right). \label{eq:denoise:shrink_r}
\end{align}
The variables $\lambda^{k+1}$ in~\eqref{eq:energy:denoise:split} and $z^{k+1}$ in~\eqref{eq:denoise:reg:MoreauYosida} are updated in the same way as in~\eqref{eq:shrink_l} and~\eqref{eq:shrink_z}.
%
The optimality condition for the update of $u^{k+1}$ reads:
\begin{align} 
0 \in - \lambda^{k+1} (f - u - r^{k+1}) + \mu \theta (u - v^k + w^k), \label{eq:denoise:condition:u}
\end{align}
leading to the following solution:
\begin{align} 
(\lambda^{k+1} + \mu \theta) u^{k+1} = \mu \theta (v^k - w^k) + \lambda^{k+1} (f - r^{k+1}). \label{eq:denoise:stepU}
\end{align}
The solutions for the update of $v^{k+1}$ and $w^{k+1}$ are obtained in the same way as in~\eqref{eq:StepV} and~\eqref{eq:stepW}, respectively.
%
%
%
\ifCLASSOPTIONcaptionsoff
  \newpage
\fi
%
%
%
{\small
\bibliographystyle{IEEETran}
\bibliography{adaptive}
}
\end{document}